\documentclass[letterpaper,oneside]{SprMixLaTeX}

\usepackage[sectionbib,longnamesfirst]{natbib}
\bibpunct{(}{)}{;}{a}{}{,}
\renewcommand{\cite}{\citep}

\usepackage{makeidx}         
\usepackage{graphicx,marvosym}        
\usepackage{multicol}        
\usepackage[bottom]{footmisc}

\usepackage{eucal,url}
\usepackage{graphics}

\usepackage{array}
\usepackage{latexsym}
\usepackage{amsmath}
\usepackage{amssymb}
\usepackage{color}
\usepackage{comment}
\usepackage[latin1]{inputenc}

\usepackage{algorithmic}
\usepackage{tabularx}
\usepackage{pbox}

\usepackage{psfrag}

\usepackage{verbatim}
\usepackage{import}

\usepackage{lscape}
\usepackage{subfigure}
\usepackage{rotating}

\usepackage{epstopdf}

\usepackage[ruled,algochapter]{algorithm2e}

\usepackage{subfigure}
\usepackage{tikz}
\usepackage{tkz-graph}
\usetikzlibrary{calc,arrows,plotmarks}

\usepackage{ifthen}

\newsavebox{\laterrefbox}
\newlength{\tempwidth}

\newcommand{\laterref}[1]{
  \sbox{\laterrefbox}{#1}%
  \ifthenelse{\lengthtest{\wd\laterrefbox > \linewidth}}
  {\setlength{\tempwidth}{\linewidth}
    \fbox{\fbox{
        \begin{minipage}{\linewidth}
          \textsl{\small #1}
        \end{minipage}
        }}
    }
  {\fbox{\fbox{#1}}}
  \ref{#1}%
}

\ifx\newtheorem\undefined
  \usepackage{amsthm}
\fi

\newlength{\defsep}
\DeclareMathOperator*{\argmin}{arg\:min}

\makeindex             

\begin{document}

\setcounter{chapter}{3}

\title{Empowerment --- An Introduction}

\author{Christoph Salge, Cornelius Glackin and Daniel Polani}

\institute{University of Hertfordshire}

\maketitle

\vspace{0.7 cm}

\section{Introduction}

Is it better for you to own a corkscrew or not? If asked, you as a human being would likely say ``yes'', but more importantly, you are somehow able to make this decision. You are able to decide this, even if your current acute problems or task do not include opening a wine bottle. Similarly, it is also unlikely that you evaluated several possible trajectories your life could take and looked at them with and without a corkscrew, and then measured your survival or reproductive fitness in each. When you, as a human cognitive agent, made this decision, you were likely relying on a behavioural ``proxy'', an internal motivation that abstracts the problem of evaluating a decision impact on your overall life, but evaluating it in regard to some simple fitness function. One example would be the idea of curiosity, urging you to act so that your experience new sensations and learn about the environment. On average, this should lead to better and richer models of the world, which give you a better chance of reaching your ultimate goals of survival and reproduction. 

But how about questions such as, would you rather be rich than poor, sick or healthy, imprisoned or free? While each options offers some interesting new experience, there seems to be a consensus that rich, healthy and free is a preferable choice. We think that all these examples, in addition to the question of tool ownership above, share a common element of preparedness. Everything else being equal it is preferable to be prepared, to keep ones options open or to be in a state where ones actions have the greatest influence on ones direct environment.    

The concept of \emph{Empowerment}, in a nutshell, is an attempt at
formalizing and quantifying these degrees of freedom (or options) that
an organism or agent has as a proxy for ``preparedness'';
preparedness, in turn, is considered a proxy for prospective fitness
via the hypothesis that preparedness would be a good indicator to
distinguish promising from less promising regions in the prospective
fitness landscape, without actually having to evaluate the full fitness landscape. 

Empowerment aims to reformulate the options or degrees of freedom that
an agent has as the agent's control over its environment; and not only
of its control --- to be reproducible, the agent needs to be aware of
its control influence and sense it. Thus, empowerment is a measure of
both the control an agent has over its environment, as well as its
ability to sense this control. Note that this already hints at two different perspectives to evaluate the empowerment of an agent. From the agent perspective empowerment can be a tool for decision making, serving as a behavioural proxy for the agent. This empowerment value can be skewed by the quality of the agent world model, so it should be more accurately described as the agent's approximation of its own empowerment, based on its world model. The actual empowerment depends both on the agent's embodiment, and the world the agent is situated in. More precisely, there is a specific empowerment value for the current state of the world (the agent's current empowerment), and there is an averaged value over all possible states of the environment, weighted by their probability (the agent's average empowerment).

Empowerment, as introduced by
\citet{klyubin2005,klyubin2005empowerment}, aims to formalize the combined
notion of an agent controlling its environment and sensing this
control in the language of information theory. The idea behind this is
that this should provide us with a utility function that is inherently
\emph{local}, \emph{universal} and \emph{task-independent}.

\begin{enumerate}
\item \textit{Local} means that the knowledge of the local dynamics of
  the agent is enough to compute it, and that it is not necessary to
  know the whole system to determine one's empowerment. Ideally,
  the information that the agent itself can acquire should be enough. 
\item \textit{Universal} means that it should be possible to apply
  empowerment ``universally'' to every possible agent-world
  interaction. This is achieved by expressing it in the language of
  information theory and thus making it applicable for any system that
  can be probabilistically expressed.

  For instance, even if an agent completely changes its morphology, it
  is still possible to compute a comparable empowerment value.
  \citet{klyubin2005empowerment} gave the examples of money in a bank
  account, of social status in a group of chimpanzees, and of sugar
  concentration around a bacterium as different scenarios, all as
  examples which would be treated uniformly by the empowerment
  formalism.
\item \textit{Task-independent} means that empowerment is not
  evaluated in regard to a specific goal or external reward state.
  Instead, empowerment is determined by the agent's embodiment in the
  world. In particular, apart from minor niche-dependent parameters,
  the empowerment formalism should have the very same structure in
  most situations.
\end{enumerate}

More concretely, the proposed formulation of empowerment defines it
via the concept of potential information flow, or channel capacity,
between an agent's actuator state at earlier times and their sensor
state at a later time. The idea behind this is that empowerment would
quantify how much an agent can reliably and perceptibly influence the
world.

\subsection{Overview}

Since its original inception by
\citet{klyubin2005,klyubin2005empowerment} in 2005, several papers
have been published about empowerment, both further developing the
formalism, and demonstrating a variety of behaviours in different
scenarios. Our aim here is to both present an overview of what has
been done so far, and to provide readers new to empowerment with an
easy entry point to the current state-of-the-art in the field. Due to the amount of content, some ideas and results are only reported in abstract form, and we would refer interested reader to the cited papers, where models and experiments are explained in greater detail. 

Throughout the text we also tried to identify the open problems and
questions that we currently see, and we put a certain emphasis on the
parameters that affect empowerment. While empowerment is defined in a
generic and general way, the review of the literature shows that there
are still several choices one can take on how to exactly apply empowerment,
and which can affect the outcome of the computation.

The remaining paper is structured as follows. First, after briefly outlining the related work previous to empowerment, we will spell out
the different empowerment hypotheses motivating the research in
empowerment. This will allow us to locate empowerment in relation to
different fields, and also makes it easier to see how and where 
insights from the empowerment formalism apply to other areas.

The next section then focusses on discrete empowerment, first, in Sec. \ref{sec:formalism} introducing the formalism, and then, in Sec. \ref{sec:discreteEx}, describing several different examples, showcasing the genericity of the approach. 

Section \ref{sec:cont-empow} then deals with  empowerment in continuous settings,
which is currently not as far developed and sees vigorous activity.
Here we will discuss the necessity for suitable approximations, and
outline the current technical challenges to provide  good but fast
approximations for empowerment in the continuous domain.

\section{Related Work} 
\label{related work}

Empowerment is based on and connects to several fields of scientific inquiry. One foundational idea for empowerment is to apply information theory to living, biological systems. \citep{gibson1979ecological} points out the importance of information in embodied cognition, and earlier work \cite{barlow1959sensory,attneave1954some} investigates the informational redundancy in an agent's sensors. Later research \cite{atick1992could} based on this identifies the importance of \textit{information bottlenecks} for the compression of redundancies, which are later formalized in information theoretic terms \cite{tishby1999}. Furthermore, it was also demonstrated that informational efficiency can be used to make sense of an agent's sensor input \cite{olsson2005sensor,lungarella2005methods}. The general trend observed in these works seems to be that nature optimizes the information processing in organisms in terms of efficiency \cite{polani2009information}.  Empowerment is, in this context, another of these efficiency principles. 

Empowerment also relies heavily on the notion that cognition has to be understood as an immediate relationship of a situated and embodied agent with its surroundings. This goes back to the ``Umwelt'' principle by \citep{von1909umwelt}, which also provides us with an early depiction of what is now commonly referred to as the perception action loop. This idea was also at the center of a paradigm shift in artificial intelligence towards \textit{enactivism} \cite{varela1992embodied,almeida2005introduction}, which postulates that the human mind organizes itself by interacting with its environment. Embodied robotics \cite{pfeifer2007body} is an approach trying to replicate these processes ``in silico''.

\subsection{Intrinsic Motivation}

Central to this body of work is the desire to understand how an
organism makes sense of the world and decides to act from its internal
perspective. Ultimately all behaviour is connected to an organism's
survival, but most natural organisms do not have the cognitive
capacity to determine this connection themselves. So, if an animal
gets burned by fire, it will not consider the fire's negative effect
on its health and potential risk of death and then move away. Instead,
it will feel pain via its sensors and react accordingly. The ability
to feel pain and act upon it is an adaptation that acts as a proxy
criterion for survival, while it still offers a certain level of
abstraction from concrete hard-wired reactions. We could say the
animal is motivated to avoid pain. Having an abstract motivation allows an agent a certain amount of adaptability; instead of acting like a stimulus-response look-up table the agent can evaluate actions in different situations according to how rewarding they are regarding its motivations.

Examining nature also reveals that not all motivations are based on
external rewards, e.g. a well-fed and pain-free agent might be driven
by an urge to explore or learn. In the following we discuss related
work covering different approaches to specify and quantify such
\textit{intrinsic motivations}. The purpose of these models is both to
better understand nature, as well as to replicate the ability of
natural organism to react to a wide range of stimuli in models for
artifical systems.

An evolution-based view of intrinsic motivations uses assumptions
about preexisting saliency sensors to generate intrinsic motivations
\cite{singh-intrinsically,singh2010intrinsically}. However, where one
does not want to assume such pre-evolved saliency sensors, one needs
to identify other criteria that can operate with unspecialized
generic sensors. 

One such family of intrinsic motivation mechanisms focusses on
evaluating the learning process. \emph{Artificial curiosity}
\cite{schmidhuber2002exploring,schmidhuber1991curious} is one of the
earlier models, where an agent receives an internal reward depending
on how ``boring'' the environment is which it currently tries to
learn. This causes the agent to avoid situations that are at either of
the extremes: fully predictable or unpredictably random.

The \emph{autotelic principle} by \citet{steels2004autotelic} tries to
formalize the concept of ``Flow'' \cite{csikszentmihalyi2000beyond}:
an agent tries to maintain a state were learning is challenging, but
not overwhelming \citep[see
also][]{gordon12:_hierar_curios_loops_activ_sensin}. Another approach
\cite{kaplan2004maximizing} aims to maximise the learning progress of
different classical learning approaches by introducing rewards for
better predictions of future states. A related idea is behind the
\emph{homeokinesis} approach, which can be considered a dynamic
version of homoeostasis. The basic principle here is to act in a way
which can be well predicted by a adaptive model of the world dynamics
\cite{der1999homeokinesis}. There is a tendency of such mechanisms to
place the agent in stable, easily predictable environments. For this
reason, to retain a significant richness of behaviours additional
provisions need to be taken so that notwithstanding the predictability
of the future, the current states carry potential for instability.

The ideas of homeokinesis are originally based on dynamical system
theory. Further studies have transferred them into the realm of
information-theoretical approaches \cite{Ay2008}. The basic idea here
is to maximise the \textit{predictive information}, the information
the past states of the world have about the future. Here, also,
predictability is desired, but predictive information will only be
large if the predictions about the future are decoded from a rich
past, which captures very similar ideas to the dynamical systems view
of homeokinesis.

The empowerment measure which is the main concept under discussion in
the present paper, also provides a universal, task-independent
motivation dynamics. However, it focusses on a different niche. It is
not designed to \emph{explore} the environment, as most of the above
measures are, but rather aims at identify preferred states in the
environment, once the local dynamics are known; if not much is known
about the environment, but empowerment is high, this is perfectly
satisfactory for the empowerment model, but not for the earlier
curiosity-based methods. Therefore, empowerment is better described as
a complement to the aforementioned methods, rather than a direct
competitor.

Empowerment has been motivated by a set of biological hypotheses, all
related to informational sensorimotor efficiency, the ability to react
to the environment and similar. However, it would be interesting to
identify whether there may be a route stemming from the underlying
physical principles which would ultimately lead to such a principle
(or a precursor thereof). For some time, the "Maximum Entropy
Production Principle" (MEPP) has been postulated as arising from first
thermodynamic principles \cite{dewar2003information,dewar2005maximum}. However, unfortunately, and according to current knowledge, the derivation from first principles still remains elusive and the current attempts at doing so unsuccessful \cite{grinstein2007comments}. If, however, one should be able to derive the MEPP from first principles, then \cite{wissner2013causal}
show that this would allow a principle to emerge on the physical
(sub-biological) level which acts as a simpler proto-empowerment which
shares to some extent several of the self-organizing properties with
empowerment, even if in a less specific way and without reference to
the ``bubble of autonomy'' which would accompany a cognitive agent.
Nevertheless, if successful, such a line may provide a route to how a
full-fledged empowerment principle could emerge from physical principles.

\section{Empowerment Hypotheses}

In this section we want to introduce the main hypotheses which motivated the development of empowerment. Neither the work presented here in this chapter, nor the work on empowerment in general is yet a conclusive argument for either of the three main hypotheses, but they should, nevertheless, be helpful to outline what empowerment can be used for, and to what different domains empowerment can be applied. Furthermore, it should also be noted, that the hypotheses are stated in a generic form which might be unsuitable for experimental testing, but this can be alleviated on a case by case basis by applying a hypothesis to a specific scenario or task. 

There are two main motivations for introducing the concept of
empowerment: one is, of course, the desire to come up with methods to
allow artificial agents to flexibly decide what to do generically,
without having a specific task designed into them in every situation. This is closely related to the idea of creating a general AI. 
The other is to search for candidate proxies of prospective fitness, which could be detected and driven towards during the lifetime of an organism to improve its future reproductive success.

From these two starting points, several implicit and explicit claims have been made about empowerment and how it would relate to phenomena in biology. In the following section we structure these claims into three main hypotheses which we would consider as driving the ``empowerment program''. This should make it easier for the reader to understand what the simulations in the later
chapters should actually demonstrate.

\subsection{Behavioural Empowerment Hypothesis}

\begin{quote}
\textit{The adaptation brought about by natural evolution produced organisms that in absence of specific goals behave as if they were maximising their empowerment.}
\end{quote}

\citet{klyubin2005,klyubin2005empowerment} argue that the direct
feedback provided by selection in evolution is relatively sparse, and
therefore it would be infeasible to assume that evolution adapts the
behaviour of organisms specifically for every possible situation.
Instead they suggest that organisms might be equipped with local,
task-independent utility detectors, which allows them to react well to
different situations. Such generic adaptation might have arisen as a
solution to a specific problem, and then persisted as a solution to
other problems, as well. This also illustrates why such a utility
function should be universal: namely, because it should be possible to
retain the essential structure of the utility model, even if the
morphology, sensor or actuators of the organism change through
evolution.

This is also based on our understanding of humans and other organisms.
We seem to be, at least in part, adapted to learn, explore and reason,
rather than to only have hard-coded reactions to specific stimuli. As
these abilities also usually generate actions, such a drive is
sometimes called \emph{intrinsic motivation}. Different approaches
have been proposed (see Sec. \ref{related work}) to formalize
motivation that would generate actions that are not caused by an
explicit external reward. Empowerment does not consider the learning
process or the agent trajectory through the world, but instead
operates as a pseudo-utility which assigns a value (its empowerment)
to each state in the world\footnote{Here we mostly adopt an
  ``objective'' perspective in that the objective states of the world
  are known and their empowerment computed. However, truly subjective
  versions of empowerment are easily definable and will be discussed
  in Sec. \ref{sec:context} as context-dependent empowerment.}.
Highly empowered states are
preferred, and the core hypothesis states that an agent or organism
attempts to reach states with high empowerment. Empowerment measures
the ability of the agent to \emph{potentially} change its future (it
does not mean that it is actually doing that). The lowest value for
empowerment is 0, which means that an agent has no influence on the
world that it can perceive. From the empowerment perspective,
vanishing empowerment is equivalent to the agent's death, and the
empowerment maximization hypothesis provides a natural drive for death
aversion.

The \emph{behaviour empowerment hypothesis} now assumes that evolution
has come up with a solution that produces similar behaviour. To
support this hypothesis, the first step would be to demonstrate that
empowerment can produce behaviour which is similar to biological
organisms in analogous situations. In turn, it should also be possible
to anticipate behaviour of biological organisms by considering how it
would affect their empowerment. If we follow this idea further and assume
that humans use empowerment-like criteria to inform their
introspection, then one would expect that those states identified by
humans as preferable would also be more likely to have high
empowerment.

For the hypothesis to be plausible, it would also be good to ensure
that empowerment is indeed local and can be computed from the
information available to the agent. Similarly, it should also be
universally applicable to different kinds of organisms; we would
expect organisms which have undergone small changes to their
sensory-motor set-up to still produce comparable empowerment values,
and for organisms that discover new modalities of interaction that this is then reflected in the empowerment landscape.

So far, we have discussed a weak version of the behavioural empowerment
hypothesis. A stronger version of the hypothesis\footnote{We do not actually put forward this stronger version for the biological realm, but mention it for completeness, and because of its relevance for empowerment in artificial agents.} would argue that an
agent actually computes empowerment. While this can be easily checked for artificial agents, in a biological
scenario, it becomes necessary to explain how empowerment could actually
be computed by the agent. The weak version of the hypothesis, instead,
says that the agents just act ``as if'' driven by empowerment, or are
using a suitable approximation. The hypothesis then states that
natural behaviours favour highly empowered behaviour
routes. 

\subsection{Evolutionary Empowerment Hypothesis}

\begin{quote}
\textit{The adaptation brought about by natural evolution increases the empowerment of the resulting organism.}
\end{quote}

Due to its universality, empowerment can in principle, be used to
compare the average empowerment of different organisms. For instance,
today, we could look at a digital organism, and then come back later
after several generations of simulated adaptation, asking whether the
organisms are now more empowered? Did that new sensor (and/or
actuator) increase the agents empowerment? The hypothesis put forward, e.g.\ by \citet{polani2009information}, is that the adaptation in nature, on average, increases an agent's empowerment. He argues that (Shannon) information operates as a ``currency of life'', which imposes an inherent cost onto an organism, and, for that reason, a well-adapted organism should have efficient information processing. On the one hand, there is some relevant information \cite{polani2006relevant} that needs to be acquired by an agent to perform at a given level, but any additional information
processing would be superfluous, and should be avoided, as it creates
unnecessary costs. Taking a look at agent morphologies, this also
means that agents should be adapted to efficiently use their sensors
and actuators. For example, a fish population living in perpetual
darkness does not have a need for highly developed eyes \cite{jeffery2005adaptive}, and it is expected that adaptation will reduce the functionality and cognitive investment (i.e.\ brain operation) related to vision. On the other hand, in the dark the detection of sound could be useful; this perceptual channel could be made even more effective by actively
generating sound that is then reflected from objects and could then be detected by the organism. The core question is: how can such potential
advantageous gradients in the space of sensorimotoric endowment be
detected?

Empowerment is the channel capacity from an agent's actuators to its
sensors, and as such, measures the efficiency of that channel. Having
actuators whose effect on the environment cannot be perceived, or
sensors which detect no change relevant to the current actions is
inefficient, and should be selected against. In short, this adaptation
would be attained by an increase of the agent's average empowerment.

A test for this hypothesis would be to evolve agents in regard to
other objectives, and then check how their empowerment develops over
the course of the simulated evolution, similar to studies about
complexity growth under evolutionary pressures \cite{yaeger2009evolution}. Another salient effect of this hypothesis would be an adaptation of an agent morphology based on empowerment should produce sensor layouts and actions which are to some degree ``sensible'' and perhaps could also be compared to those found in nature.

\subsection{AI Empowerment Hypothesis}

\begin{quote}
\textit{Empowerment provides a task-independent motivation that generate AI behaviour which is beneficial for a range of goal-oriented behaviour.} 
\end{quote}
In existing work, it was demonstrated that empowerment can address
quite a selection of AI problems successfully (see the remaining chapter for a selection); amongst these are pole balancing,
maze centrality and others. However, a clear contraindication exists
for its use: if an \emph{externally} desired goal state is not highly
empowered, then an empowerment-maximising algorithm is not going to
seek it out. Opposed to that, such tasks are the standard domain of
operation for traditional AI algorithms.

However, in the realm or robotics there have been developments to design robots that are not driven by specific goals, but motivated by exploring
their own morphology or other forms of intrinsic motivation. The idea
is to build robots that learn and explore, rather than engineer
solutions for specific problems determined externally and in advance.
Here, empowerment offers itself as another alternative. While
empowerment is not designed to explicitly favour exploration, it has
an inbuilt incentive to avoid behaviour that leads to a robot being
stuck. Having no options available to an agent is bad for empowerment. 
Non-robotic AI could also benefit from this approach, but since
empowerment is defined over the agent world dynamics, there needs to
be a clear interface between an agent and the world over which it can
be computed: in this case, there needs to be some kind of substitute
for embodiment or situatedness. On the other hand, for the robotics
domain it is also important that empowerment can be computed in real
time and be applied to continuous variables.

The concrete and relevant question would be under which circumstances
empowerment would provide a good solution, both in robotic and
non-robotic settings? Furthermore, in what situations would maximising
empowerment be helpful for a later to be specified task? To approach
this question it is helpful to apply empowerment to a wider range of
AI problems and inspect its operation in the different scenarios. The
remaining chapter will showcase several such examples and discuss the
insights gained from these.

In the robotic domain, one faces additional challenges, most
prominently the necessity to handle empowerment in continuous spaces.
This is discussed in Sec.~\ref{sec:cont-empow}. Note, however, that
there is still very little current experience on deploying empowerment
on real robots, with exception of a basic proof-of-principle
context reconstruction example on an AIBO robot \cite{klyubin2008keep}.

\section{Formalism}
\label{sec:formalism}

Empowerment is formalized using terms from information theory, first
introduced by Shannon \cite{Shannon1948}. To define a consistent
notation, we begin by introducing several standard notions. Entropy is
defined as
\begin{equation}
\label{entropy}
H(X) = - \sum_{x \in \mathcal{X}} p(x) \log p(x)
\end{equation}
where $X$ is a discrete random variable with values $x \in
\mathcal{X}$, and $p(x)$ is the probability mass function such that
$p(x) = Pr \{ X = x \}$. Throughout this paper base 2 logarithms are
used by convention, and therefore the resulting units are in
\textit{bits}. Entropy can be understood as a quantification of
uncertainty about the outcome of $X$ before it is observed, or as the
average surprise at the observation of $X$. Introducing another random variable $Y$ jointly distributed with $X$, enables the definition of \textit{conditional entropy} as
\begin{equation}
\label{conditionalEntropy}
H(X|Y) = - \sum_{x \in \mathcal{X}} p(y) \sum_{y \in \mathcal{Y}} p(x|y) \log p(x|y).
\end{equation}
This measures the remaining uncertainty about $X$ when $Y$ is known.
Since Eq.~(\ref{entropy}) is the general uncertainty of $X$, and
Eq.~(\ref{conditionalEntropy}), is the remaining uncertainty once $Y$ has been observed, their difference, called \textit{mutual information}, quantifies the average information one can gain about $X$ by observing $Y$. Mutual information is defined as
\begin{equation}
\label{mutualInformation}
I(X;Y) = H(Y) - H(Y|X).
\end{equation}
The mutual information is symmetric (see \cite{Cover1991}), and it holds that
\begin{equation}
\label{mutualInformationSymmetry}
I(X;Y) = H(Y) - H(Y|X) = H(X) - H(X|Y).
\end{equation}
Finally, a quantity which is used in communication over a noisy
channel to determine the maximum information rate that can be reliably
transmitted, is given by the \emph{channel capacity}:
\begin{equation}
\label{channelCapacity}
C = \max_{p(x)} I(X;Y)\;.
\end{equation}
These concepts are fundamental measures in classical information theory.

Now, for the purpose of formalizing empowerment, we will now
reinterpret the latter quantity in a causal context, and specialize
the channel we are considering to the actuation-perception channel.

\subsection{The Causal Interpretation of Empowerment}

Core to the empowerment formalism is now the potential \emph{causal}
influence of one variable (or set of variables: the actuators) on
another variable (or set of variables: the sensors). Further below, we
will define the framework to define this in full generality; for now,
we just state that we need to quantify the potential \emph{causal effect}
that one variable has on the other.

When we speak about causal effect, we specifically consider the
interventionist notion of causality in the sense of \citet{Pearl2000}
and the notion of causal information flow based upon it \cite{Ay2008a}.
We sketch this principle very briefly and refer the reader to the
original literature for details. 

To determine the causal information flow $\Phi(X \rightarrow Y)$ one
cannot simply consider the observed distribution $p(x,y)$, but has to
probe the distribution by actively intervening in $X$. The change
resulting from the intervention in $X$ (which we denote by $\hat{X}$)
is then observed in the system and used to construct the
interventional conditional $p(y|\hat{x})$. This interventional
condition will then be used as the causal channel of interest. While
(causal) information flow according to \cite{Ay2008a} has been defined
as the mutual information over that channel for an independent
interventional input distribution, empowerment considers the
\emph{maximal} potential information flow, i.e.\ it is not based on
the actual distribution of the input variable $X$ (with or without
intervention), but considers the maximal information flow that could
possibly be induced by a suitable choice of $X$. This, however, is
nothing other than the channel capacity
\begin{equation}
\label{channelCapacity}
C(X \rightarrow Y) = \max_{p(\Hat x)} I(\Hat X;Y).
\end{equation}
for the channel defined by  $p(y|\hat{x})$, where by the hat we
indicate that this is a channel where we intervene in $X$.

There is a well-developed literature on how to determine the conditional probability distribution
$p(y|\hat{x})$ necessary to compute empowerment, for some approaches,
see \cite{Pearl2000,Ay2008a}. This interventional conditional probability distribution can then be treated as the channel; and the channel capacity, or empowerment, can be computed with established methods, such as the Blahut-Arimoto algorithm \cite{blahut1972,arimoto1972}. 

For the present discussion, it shall suffice to say that empowerment
can be computed from the conditional probability distribution of
observed actuation/sensing data, as long as we can ensure that the
channel is a causal pair, meaning we can rule out any common cause,
and any reverse influence from $y$ onto $x$.

\subsection{Empowerment in the Perception Action Loop}
\label{sec:empow-perc-acti}
The basic idea behind empowerment is to measure the influence of an
agent on its environment, and how much of this influence can be
perceived by the agent. In analogy to control theory, it is
essentially a combined measure of controllability (influence on the
world) and observability (perception by the agent), but, unlike in the
control-theoretic context, where controllability and observability
denote the dimensionality of the respective vector spaces or
manifolds, empowerment is a fully information-theoretic quantity: This
has two consequences: the values it can assume are not confined to
integer dimensionalities, but can range over the continuum of
non-negative real numbers; and, secondly, it is not limited to linear
subspaces or even manifolds, but can, in principle, be used in all
spaces for which one can define a probability mass measure.

\begin{figure}
\begin{center}
\subfigure[\label{fig:pal1a}Memoryless Perception Action Loop]{
\begin{tikzpicture}[node distance=1.2cm]
  \GraphInit[vstyle=Dijkstra]
  \SetVertexMath \renewcommand*{\VertexLineColor}{white}
  
  \Vertex[L=R_{t-1},Lpos=90]{R0}
  \SOEA[L=S_{t-1},Lpos=-90](R0){S0}
  \EA[L=A_{t-1},Lpos=-90](S0){A0}
  \NOEA[L=R_{t},Lpos=90](A0){R1}
  \SOEA[L=S_{t},Lpos=-90](R1){S1}
  \EA[L=A_{t},Lpos=-90](S1){A1}
  \NOEA[L=R_{t+1},Lpos=90](A1){R2}
	
  \WE[empty](R0){E0}
  \SOWE[empty](R0){E1}
  \EA[empty](R2){E2}
  \SOEA[empty](R2){E3}
	
  \SetUpEdge[style={post}]
  \tikzstyle{EdgeStyle}=[line width=.5pt]
  \tikzstyle{LabelStyle}=[left=3pt]
  \tikzstyle{LabelStyle}=[above=3pt]
  \Edge(R0)(R1)
  \Edge(R1)(R2)
  \Edge(R0)(S0)
  \Edge(S0)(A0)
  \Edge(A0)(R1)
  \Edge(R1)(S1)
  \Edge(S1)(A1)
  \Edge(A1)(R2)
	
  \SetUpEdge[style={post}, color=gray]
  \tikzstyle{EdgeStyle} = [line width=.5pt,dashed]
  \Edge(E0)(R0)
  \Edge(E1)(R0)
  \Edge(R2)(E2)
  \Edge(R2)(E3)
\end{tikzpicture}
}
\subfigure[\label{fig:pal1b}Perception Action Loop with Memory]{
\begin{tikzpicture}[node distance=1.2cm]
  \GraphInit[vstyle=Dijkstra]
  \SetVertexMath \renewcommand*{\VertexLineColor}{white}
  
  \Vertex[L=R_{t-1},Lpos=90]{R0}
  \SOEA[L=S_{t-1},Lpos=-90](R0){S0}
  \SOEA[L=M_{t-1},Lpos=-90](S0){M0}
  \NOEA[L=A_{t-1},Lpos=-90](M0){A0}
  \NOEA[L=R_{t},Lpos=90](A0){R1}
  \SOEA[L=S_{t},Lpos=-90](R1){S1}
  \SOEA[L=M_{t},Lpos=-90](S1){M1}
  \NOEA[L=A_{t},Lpos=-90](M1){A1}
  \NOEA[L=R_{t+1},Lpos=90](A1){R2}
	
  \WE[empty](R0){E0}
  \SOWE[empty](R0){E1}
  \EA[empty](R2){E2}
  \SOEA[empty](R2){E3}
  \WE[empty](M0){X0}
  \EA[empty](M1){X1}
	
  \SetUpEdge[style={post}]
  \tikzstyle{EdgeStyle}=[line width=.5pt]
  \tikzstyle{LabelStyle}=[left=3pt]
  \tikzstyle{LabelStyle}=[above=3pt]
  \Edge(R0)(R1)
  \Edge(R1)(R2)
  \Edge(R0)(S0)
  \Edge(S0)(A0)
  \Edge(A0)(R1)
  \Edge(R1)(S1)
  \Edge(S1)(A1)
  \Edge(A1)(R2)
  \Edge(M0)(M1)
\Edge(S0)(M0)	
\Edge(M0)(A0)	
\Edge(S1)(M1)	
\Edge(M1)(A1)
\Edge(S0)(M0)

  \SetUpEdge[style={post}, color=gray]
  \tikzstyle{EdgeStyle} = [line width=.5pt,dashed]
  \Edge(E0)(R0)
  \Edge(E1)(R0)
  \Edge(R2)(E2)
  \Edge(R2)(E3)
  \Edge(X0)(M0)	
\Edge(M1)(X1)
\end{tikzpicture}
}
\end{center}
\caption{Causal Bayesian network of the perception-action loop,
unrolled in time, showing (a) a memoryless model, (b) a model including agent memory. In the memoryless model the agent's actions $A_t$ only depend on its current sensor inputs $S_t$, while the perception action loop with memory allows for agent models in which the agent can store information from sensor inputs in the past in $M$, and use this information later for its decision making in $A$.}
\label{fig:pal1}
\end{figure}
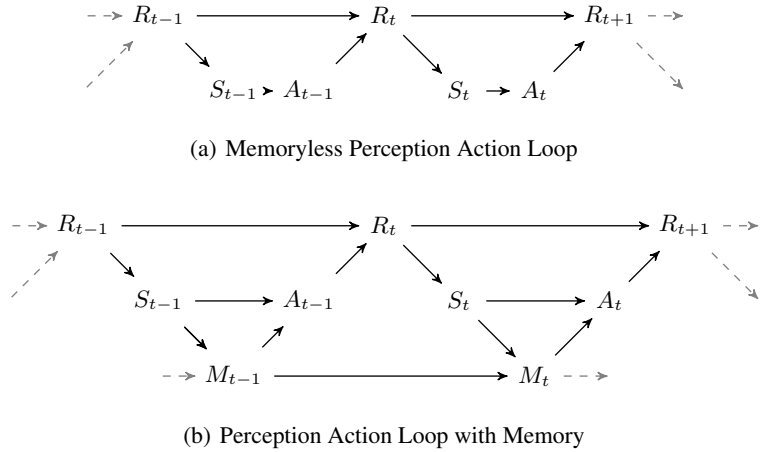

We formalize the concept of empowerment, as stated earlier, as the
channel capacity between an agent's actions at a number of times and
its sensoric stimuli at a later time. To understand this in detail,
let us first take a step back and see how to model an agent's
interaction with the environment as a causal Bayesian network (CBN).
In general we are looking at a time-discrete model where an agent
interacts with the world. This can be expressed as a perception-action
loop, where an agent chooses an action for the next time step based on
its sensor input in the current time step. This influences the state
of the world (in the next time step), which in turn influences the
sensor input of the agent at that time step. The cycle then repeats
itself, with the agent choosing another action. Note that this choice
of action might also be influenced by some internal state of the agent
which carries information about the agent's past. To model this, we
define the following four random variables:
\begin{description}
\item[$A$:] the agent's actuator\footnote{Saying \emph{actuator}
    implicitly includes the case of multiple actuators. In fact, it is
    the most general case. Multiple actuators (which can be
    independent of each other) can always be written as being
    incorporated into one single actuator variable.}
  which takes values $a \in \mathcal{A}$
 \item[$S$:] the agent's sensor which takes values $s \in \mathcal{S}$
 \item[$M$:] the agent's internal state (or memory) which takes values $m \in \mathcal{M}$
 \item[$R$:] the state of the environment which takes values $r \in \mathcal{R}$
\end{description}
Their relationship can be expressed as a time-unrolled CBN, as seen in
Fig.~\ref{fig:pal1a}.
 
Empowerment is then defined as the channel capacity between the
agent's actuators $A$ and its own sensors $S$ at a later point in
time, here, for simplicity, we assume the next time step:
\begin{equation}
\mathfrak{E} := C(A_t\rightarrow S_{t+1}) \equiv \max_{p(a_t)} I(S_{t+1};A_t)\;.
\label{eq:empowerment}
\end{equation}
This is the general empowerment of the agent. In the following text we
will use $\mathfrak{E}$ as a shorthand for the causal channel capacity
from the sensors to the actuators.\label{sec:general-empowerment}

Note that the maximization implies that it is calculated under the
assumption that the controller which chooses the action $A$ is free to
act, and is not bound by possible behaviour strategy $p(a|s,m)$.
Importantly, the distribution $p^*(a)$ that achieves the channel
capacity is different from the one that defines the actions of an
empowerment-driven agent. Empowerment considers only the
\emph{potential} information flow, so the agent will only calculate
how it \textit{could} affect the world, rather than actually carry out
its potential.

\subsection{$n$-step empowerment}

In Sec.~\ref{sec:empow-perc-acti}, we considered empowerment as a
consequence of a single action taken and the sensor being read out in
the subsequent state. However, empowerment, as a measure of the
sensorimotor efficiency, may start distinguishing the characteristics
of the agent-environment interaction only after several steps.
Therefore, a common generalization of the concept is the
\textit{$n$-step} empowerment. In this case we consider not a single
action variable, but actually a sequence of action variables for the
next $n$ time steps:
$(A_{t+1},\dots,A_{t+n})$. We we will sometimes condense these into a
single action variable $A$ for notational convenience.
The sensor variable is the resulting sensor state in the following
timestep $S_{t+n+1}$, again sometimes denoted by $S$. Though it is not
the most general treatment possible, here we will consider only
``open-loop'' action sequences, i.e.\ action sequences which are
selected in advance and then carried out without referring to a sensor
observation until the final observation $S_{t+n+1}$. This drastically
simplifies both computations and theoretical considerations, as the
different possible action sequences $A$ can be treated as if they were
separate atomic actions with no inner structure\footnote{Future work
  will investigate the effect of feedback, i.e.\ closed-loop
  sequences. However, the current hypothesis is that there will be
  little qualitative and quantitative difference for most scenarios,
  with significantly increased computational complexity.}.

As mentioned $A$ can typically contain actuator variables from several
time steps and can also incorporate several variables per time step.
$S$ is typically chosen to contain variables that are strictly
temporally downstream from all variables in $A$, to ensure a clean
causal interpretation of the effect of $A$ on $S$. However, the less
studied concept of \textit{interleaved empowerment} has been mentioned
in \cite{klyubin2008keep}, where $S$ contains sensor variables that
lie before some variables in $A$\footnote{The interpretation of
  interleaved empowerment is slightly subtle and still subject to
  study, as in this case $S$ is then capturing rather an aspect of the
  richness of the action sequences and the corresponding action
  history, in addition to the state dynamics of the system.}.

\subsection{Context-dependent Empowerment}
\label{sec:context}

Until now, we have considered empowerment as a generic characterization of
the information efficiency of the perception-action loop. Now we go a
step further and resolve this informational efficiency in more detail;
specifically, we are going to consider empowerment when the system
(e.g.\ agent and environment) is in different states. Assuming that the
state of the system is given by $r$, it will in general affect the
effect of the actions on the later sensor states, so that one now
considers $p(s|a,r)$ and defines empowerment for the world being in
state $r$ as
\begin{equation}
\mathfrak{E}(r)=\max_{p(a)}I(S;A|r),
\label{eq:context}
\end{equation} 
which is referred to as \emph{state-dependent empowerment}. This also
allows us to define the average state-dependent empowerment for an
agent that knows what state the world is in as
\begin{equation}
\mathfrak{E}(R)=\sum_{r\in R} p(r)\mathfrak{E}(r)
\end{equation}
Note that this is different from the general empowerment: the general 
empowerment in Sec.~\ref{sec:general-empowerment} does not distinguish
between different states. If different perception-action loop
characteristics $p(s|a)$ are not resolved, the general empowerment
can be vanishing, while average state-dependent empowerment is
non-zero. In other words, empowerment can depend on being able to
resolve states which affect the actuation-sensing channel.

In general, an agent will not be able to resolve all states in the
environment, and will operate using a limited sensoric resolution of
the world state. When we assume this, the agent might still be able to
recognize that the world is in a certain context $k \in K$, based on
memory and sensor input. So, an agent might not know its precise state
in the world, but may be able to identify some coarse position, e.g.\
that it might be north or south of some distinct location.
\citet{klyubin2008keep} demonstrate an example of how such a context
can be created from data. Based on this context, it is then possible
to define the marginal conditional distribution $p(s|a,k)$, which then
allows us to compute the (averaged) contextual empowerment for $K$ as
\begin{equation}
\mathfrak{E}(K)=\sum_{k\in K} p(k)\mathfrak{E}(k)
\end{equation}    
In comparison, context free empowerment $\mathfrak{E}_{\text{free}}$
has no assumption about the world and is based on the marginal
distribution $p(s|a) = \sum_r p(s|a,r)p(r)$ of all world states. This
is the empowerment that an agent would calculate which has no
information about the current world state. It can be shown
\cite{capdepuy2010} with Jensen's Inequality that
\begin{equation}
\mathfrak{E}_{\text{free}} \leq \mathfrak{E}(K) \leq \mathfrak{E}(R)
\end{equation}
This implies \citep[see also][]{klyubin2008keep} that there is a (not
necessarily unique) minimal optimal context $K_{\text{opt}}$ that best characterizes
the world in relation to how the agent's actions affect the world,
defined by:
\begin{equation}
K_{\text{opt}} = \argmin\limits_{\substack{K\\ \mathfrak{E}(K) = \mathfrak{E}(R)}} H(K).
\end{equation}
Such a context $K_{\text{opt}}$ is one which leads to the maximal
increase in contextual empowerment. \citet{klyubin2008keep} argues
that such an agent internal measure could be useful to develop
internal contexts which are purely intrinsic and based on the agent
sensory-motor capacity, and thereby allow developing an understanding
of the world based on the way they are able to interact with it.

\subsection{Open vs. Closed-Loop Empowerment}

An important distinction to make is the one between open- and closed-loop
empowerment. Open-loop empowerment treats the perception-action loop
like a unidirectional communication channel, and assumes that all
inputs are chosen ahead of time and without getting any feedback about
their source. Closed-loop empowerment is computed under the assumption
that some of the later actions in $n$-step empowerment can change in
reaction to the current sensor state.

In most of the existing work, empowerment calculations have been
performed with open-loop empowerment only. The framework for this
simplest of cases of communication theory is well developed and long
known. For the more intricate cases using feedback,
\citet{capdepuy2010} pointed out that \textit{directed
  information}~\cite{massey90:_causal_feedb_and_direc_infor} could be used to simplify
the computation of closed loop empowerment, and demonstrated for an
example how feedback increases empowerment.

\subsection{Discrete Deterministic Empowerment}
\label{sec:discr-determ-empow}

A \textit{deterministic} world is one where each action leads to one
specific outcome, i.e.\ for every $a \in \mathcal{A}$ there is exactly
one $s_a \in \mathcal{S}$ with the property that
\begin{equation}
p(s|a) = 
\begin{cases}
1 \ \text{if} \ s=s_a \\
0 \ \text{else}\;.
\end{cases}
\end{equation}
Since here every action only has one outcome, it is clear that the
conditional uncertainty of $S$ given $A$ is zero, i.e., $H(S|A) = 0$.
From Eq.~(\ref{mutualInformationSymmetry}) it follows then that
\begin{equation}
\mathfrak{E} = \max_{p(a)}(A;S)= \max_{p(a)} H(S).
\end{equation}
Since the entropy is maximal for a uniform distribution, $S$ can be
maximised by choosing any input distribution $p(a)$ which results in a
uniform distribution over the set of all reachable states of $S$, i.e\
over the set $\mathcal{S}_{\mathcal{A}} = \lbrace s \in \mathcal{S}
\vert \exists a \in \mathcal{A}:p(s|a) \geq 0 \rbrace $. As a result, empowerment for the discrete deterministic case reduces
to
\begin{equation}
\label{eq:1}
\mathfrak{E} = - \sum_{s \in \mathcal{S}_{\mathcal{A}}} \frac{1}{|\mathcal{S}_{\mathcal{A}}|} \log{( \frac{1}{|\mathcal{S}_{\mathcal{A}}|}) } = \log(|\mathcal{S}_{\mathcal{A}}|).
\end{equation}
The bottom line is that in a discrete deterministic world empowerment
reduces to the logarithm of the number of sensor states reachable with
the available actions. This means empowerment, in the deterministic case, is fully determined by how many distinguishable states the agent can reach.

\subsection{Non-deterministic Empowerment Calculation}

If noise is present in the system, an action sequence $a$ will lead to
multiple outcomes $s$, and thus, we have to consider an actual output
distribution $p(s|a)$. In this case, the optimizing distribution needs
to be determined using the standard Blahut-Arimoto (BA)
algorithm \cite{blahut1972,arimoto1972} which is an expectation
maximization-type algorithm for computing the channel capacity.

BA iterates over distributions $p_{k}(\vec{a})$, where $k$ is the
iteration counter, converging towards the distribution that maximises
channel capacity, and thereby towards the empowerment value defined in
Eq.~(\ref{eq:context}). Since the action variable $A$ is discrete and
finite we are dealing with a finite number of actions $a_v \in
\mathcal{A}$, with $v = 1, ..., n$. Therefore $p_{k}(\vec{a})$ in the
$k$-th iteration can be compactly represented by a vector
$p_{k}(\vec{a}) \equiv (p_{k}^{1},...,p_{k}^{n})$, with $p_{k}^{v}
\equiv \Pr(A=a_v)$, the probability that the action $A$ attains the
value $a_v$. Furthermore, let $s \in \mathcal{S}$ be the possible
future states of the sensor input as a result of selecting the various
actions with respect to which empowerment is being calculated, and $r
\in \mathcal{R}$ is the current state of the environment.
If we assume that $S$ is continuous we can follow the general outline from \cite{jung2011empowerment}, and define, for
notational convenience, the variable $d_{v,k}$ as:
\begin{equation}
\label{dvk}
d_{v,k} := \int_{\mathcal{S}} p(s|r,\vec{a}_{v} ) \log \left[  \frac{p(s|r,\vec{a}_{v})}{\sum^{n}_{i=1} p(s | r, \vec{a}_{i}) p^{i}_{k}} \right] d s.
\end{equation}
While this is the more general case, this integral is difficult to
evaluate for arbitrary distributions of $S$. We will later discuss, in
Sec. \ref{JMCI}, how this integral can be approximated, but even
the approximations are very computationally expensive. If we are
dealing with discrete and finite $S$ we can simply define $d_{v,k}$
with a sum as
\begin{equation}
d_{v,k} := \sum_{s \in \mathcal{S}} p(s|r,\vec{a}_{v} ) \log \left[  \frac{p(s|r,\vec{a}_{v})}{\sum^{n}_{i=1} p(s | r, \vec{a}_{i}) p^{i}_{k}} \right].
\end{equation}
The definition of $d_{v,k}$ encapsulates the differences between a continuous and discrete $S$. Therefore, the following parts of the BA algorithm are identical for both cases. The BA begins with initialising
$p_{0}(\vec{a})$ to be (e.g.) uniformly distributed\footnote{In
  principle, any distribution can be selected, provided none of the
  initial probabilities is 0, as the BA-algorithm cannot turn a
  vanishing probability into a finite one.}, by simply setting
$p_{0}^{v} = \frac{1}{n}$ for all actions $v = 1,...,n$. At each
iteration $k \geq 1$, the new approximation for the probability
distribution $p_{k}(\vec{a})$ is obtained from the old one
$p_{k-1}(\vec{a})$ using
\begin{equation}
\label{pk}
p_{k}^{v} := \frac{1}{z}_{k} p_{k-1}^{v} \exp (d_{v,k-1})
\end{equation}
where $z_{k}$ is a normalisation parameter ensuring that the approximation for the probability distribution $p_{k}(\vec{a})$ sum to one for all actions $v = 1,...,n$, and is defined as
\begin{equation}
\label{zk}
z_{k} := \sum^{n}_{v=1} p_{k-1}^{v} \exp (d_{v,k-1}).
\end{equation}
Thus $p_{k}(\vec{a})$ is calculated for iteration step $k$, it can then be used to obtain an estimate $\mathfrak{E}_{k}(r)$ for the empowerment $\mathfrak{E}(r)$ using
\begin{equation}
\label{EmpowermentEstimate}
\mathfrak{E}_{k}(r) = \sum_{v=1}^{n} p^{v}_{k} \cdot d_{v,k}.
\end{equation}
The algorithm can be iterated over a fixed number of times or until the absolute difference $|\mathfrak{E}_{k}(r) - \mathfrak{E}_{k-1}(r)|$ drops below an arbitrary chosen threshold $\epsilon$.




\section{Discrete Examples}
\label{sec:discreteEx}

\subsection{Maze}

Historically, the first scenario used to illustrate the properties of
empowerment was a maze setting introduced in \cite{klyubin2005}. Here,
the agent is located in a two-dimensional grid world. The agent has
five different actions; it can move to the adjacent squares north,
east, south and west of it, or do nothing. An outer boundary and
internal walls block the agents movement. If an agent chooses the
action to move against a wall, it will not move.

The states of the agent's action variable $A$ for n-step empowerment
are constituted by all $5^n$ action sequences that contain $n$
consecutive actions. The resulting sensor value $S$ consists of the
location of the agent at time step $t_{n+1}$, after the last action
was executed. Since we are dealing with a discrete and deterministic
world, empowerment can be calculated as in Eq.~\eqref{eq:1} in
Sec.~\ref{sec:discr-determ-empow} by taking the logarithm of all
states reachable in $n$ steps.

\begin{figure}[h]
\centerline{\includegraphics[width=3.0in]{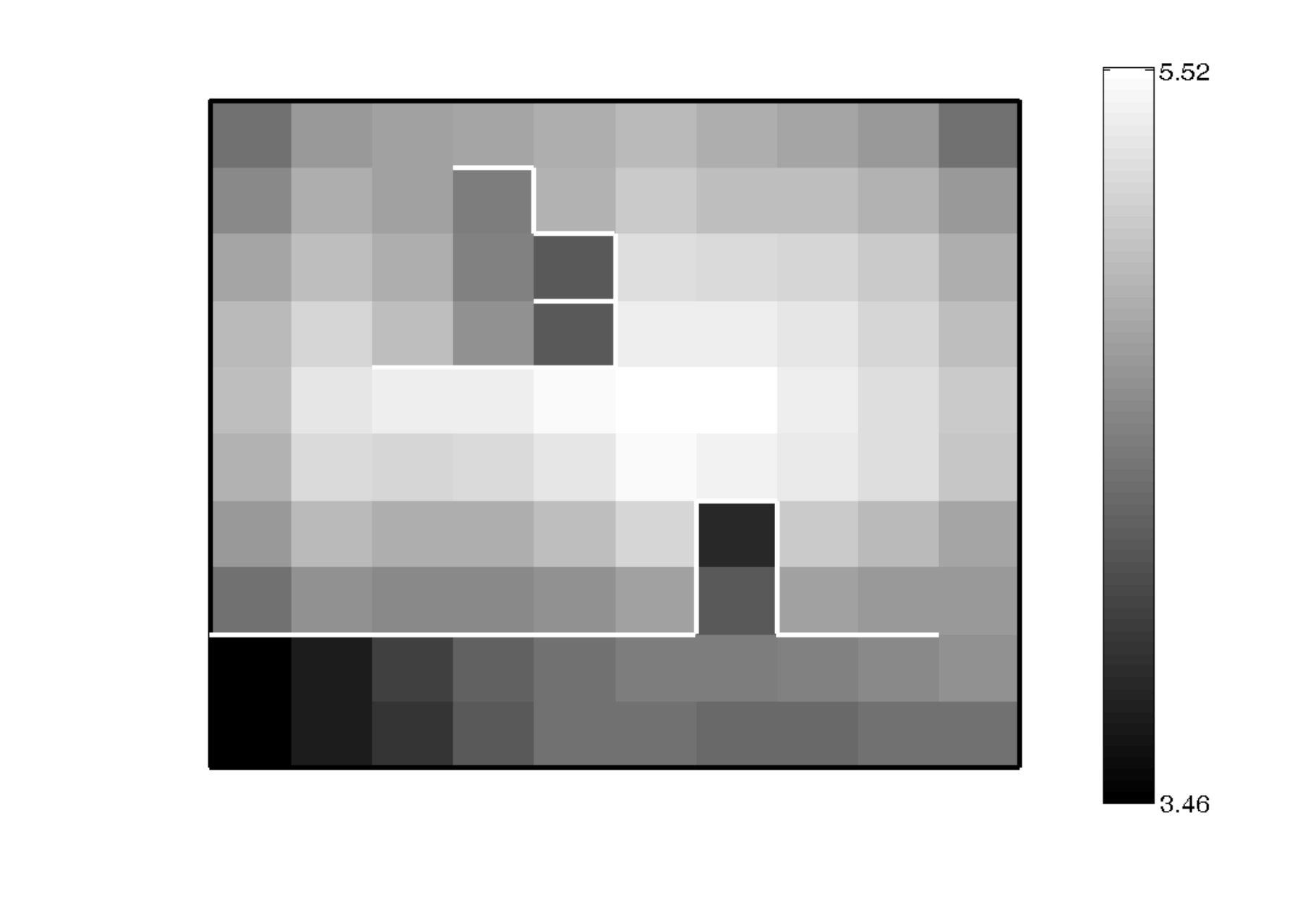}}
\caption{The graph depicts the empowerment values for 5 step action
  sequences for the different positions in a 10 $\times$ 10 maze.
  Walls are shown in white, and cells are shaded according to
  empowerment. As the key suggests empowerment values are in the range
  [3.46, 5.52] bits. This figure demonstrates that by simply assessing
  its options (in terms of movement possibilities) reflected in its
  empowerment, the agent can discover various features of the world.
  The most empowered cells in the labyrinth are those that can
  reliable reach the most positions within the next 5 steps. The graph
  is a reproduction of the results in \cite{klyubin2005}}
\label{Labyrinth}
\end{figure}

\subsection{Average Distance vs. Empowerment}

In this maze example, empowerment is directly related to how many
states an agent can reach within the next $n$ steps. Now, note that,
via the agent's actions, a Finsler metric-like
\cite{wilkens95:_finsl_geomet_low_dimen_contr_theor,Lopez:2000:SMA:587435.587483}\label{finsler-note} structure is implied on the maze, namely the minimum number of action
steps necessary to move from one given position in the maze to a
target position. Calculating $n$-step empowerment for the current
location in the maze then is simply the logarithm of all states with a
distance of $n$ or less to the current state.

Although this $n$-step horizon provides empowerment with an
essentially local ``cone of view'', \citet{klyubin2005} showed in the
maze example that empowerment of a location is negatively correlated
with the average distance of that location to \emph{all} other
locations in the maze. The first is a local, the latter, however, a
global property. This indicates that the local property of $n$-step
reachability (essentially $n$-step empowerment) would relate to a global
property, namely that of average distance. 

It is a current study objective to which extent this local/global
relation might be true, and under which conditions. Wherever it
applies, the empowerment of an agent (which can be determined from
knowledge of the local dynamics, i.e.\ how are my next $n$-steps going
to affect the world) could then be used as a proxy for certain
global properties of the world, such as the average distance to all
other states. It is clear that this cannot, in general, be true, as
outside of the empowerment horizon $n$, an environment could change
its characteristics drastically, unseen to the ``cone of view'' of the
agent's local empowerment. However, many relevant scenarios have some
regularity pervading the whole system which has the opportunity to be
detected by empowerment.

This motif was further investigated by \citet{anthony2008preferred},
who studied in more detail the relationship between graph centrality
and empowerment. The first chosen model was a two-dimensional grid
world that contained a pushable box, similar to \cite{klyubin2005}.
The agent could take five actions; move north, south, west, east, or
do nothing. If the agent moves into the location with the box,
the box would be pushed into the next square. The state space, the set
of possible world configurations, included the position of the agent,
and also the position of the box.

The complete system can be modelled as a directed labelled graph,
where each node represents a different state of the world and the
directed edges, labelled with actions, represent the transitions from
one state to another under a specific action. For an agent with 5
possible actions, all nodes have 5 edges leading away from them. This
is a generic representation of any discrete and deterministic model.
The advantage of this representation is that it provides a core
characterization of the system in graph-theoretic language which is
abstracted away from a physical representation. As before, the
distance from one state to another depends on how many actions an
agent needs to move from the first to the second state. In general,
this defines a Finsler metric-like structure (see
Sec.~\ref{finsler-note}), and is not necessarily tied to physical
distance.

\citet{anthony2008preferred} then studied the correlation between
closeness centrality and empowerment, both for the previously
described box pushing scenario. In addition, he considered a different
scenario, namely scale-free random networks as transition graphs. As
before, one can consider closeness centrality (which is a global
property), and empowerment (which can be calculated from a local
subset of the graph). \citet{anthony2008preferred} find that: 
\begin{quote}
  ``these results show a strong indication of certain global aspects of
  various worlds being `coded' at a local level, and an appropriate
  sensory configuration can not only detect this information, but can
  also use it\dots ''
\end{quote}
It is, however, currently unknown how generally and under which
circumstances this observation holds. As mentioned before, it is
possible to construct counterexamples. A natural example is the one
that Anthony et al. note in their discussion, namely that the
relationship breaks down for the box pushing example when the agents
horizon does not extend to the box; in this case, the agent is too far
away for $n$-step empowerment to be affected by the box. This might
indicate that a certain degree of structural homogeneity throughout
the world is necessary for this
relation to hold, and that the existence of different ``pockets'' in the
state space with different local rules would limit the ability of
empowerment to estimate global properties. After all, if there is
a part of the world that is radically different from the one the agent
is in, and the agent is not able to observe it in the near future, the
current situation may not be able to be informative concerning that
remote part of the world.

At present, however, it remains an open question how empowerment
relates to global properties, such as in the example of graph
centrality or average distance. No full or even partial
characterization of scenarios where empowerment correlates to
global values is currently known.

\subsection{Sensor and Actuator Selection}
\label{boxpushing}

\begin{figure}
\centering
\begin{minipage}[tbh]{0.15\columnwidth}
\hfill
\end{minipage}
\begin{minipage}[tbh]{0.25\columnwidth}
\centering
stationary~box
\end{minipage}
\begin{minipage}[tbh]{0.25\columnwidth}
\centering
pushable~box
\end{minipage}
\vfill
\vspace{0.25em}
\begin{minipage}[tbh]{0.15\columnwidth}
the~agent\\
does~not\\
perceive\\
the~box
\end{minipage}
\begin{minipage}[tbh]{0.25\columnwidth}
\centering
\fbox{\includegraphics[width=0.8\textwidth]{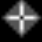}}
\vfill
\vspace{0.25em}
\textbf{a.}~~~$\mathfrak{E} \in [5.86; 5.93]$\\~~~
\end{minipage}
\begin{minipage}[tbh]{0.25\columnwidth}
\centering
\fbox{\includegraphics[width=0.8\textwidth]{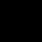}}
\vfill
\vspace{0.25em}
\textbf{b.}~~~$\mathfrak{E} = \log_{2}61$~~~~~\\~~~~~~$\approx 5.93$
bit
\end{minipage}
\vfill
\begin{minipage}[tbh]{0.76\columnwidth}
\vspace{0.5em}
\hfill
\hrule
\vspace{0.5em}
\end{minipage}
\vfill
\begin{minipage}[tbh]{0.15\columnwidth}
the~agent\\
perceives\\
the~box
\end{minipage}
\begin{minipage}[tbh]{0.25\columnwidth}
\centering
\fbox{\includegraphics[width=0.8\textwidth]{box-map-nopush-p}}
\vfill
\vspace{0.25em}
\textbf{c.}~~~$\mathfrak{E} \in [5.86; 5.93]$
\end{minipage}
\begin{minipage}[tbh]{0.25\columnwidth}
\centering
\fbox{\includegraphics[width=0.8\textwidth]{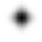}}
\vfill
\vspace{0.25em}
\textbf{d.}~~~$\mathfrak{E} \in [5.93; 7.79]$
\end{minipage}

\caption{Empowerment maps for 5-step empowerment in a 2 dimensional grid world, containing a box in the center. The scenarios differ by whether the box can be pushed by the agent or not, and whether the agent can perceive the box. Black indicates the highest empowerment. Figure reproduced from \cite{klyubin2005}}
\label{box}
\end{figure}

An agent's empowerment is not only affected by the state of the world,
i.e., the context of the agent, but also depends on what the agent's
sensors and actions are. This was illustrated by \citet{klyubin2005}
by variation of the previously mentioned box-pushing example. In all
scenarios we are dealing with a two dimensional grid world where the
agent has five different actions. The center of the world contains a
box. In Fig.~\ref{box} we see the 5-step empowerment values for the
agent's starting position in four different scenarios. The scenarios
differ depending on
\begin{enumerate}
\item whether the agent can perceive the box and
\item whether the agent can push the box.
\end{enumerate}

In Fig~\ref{box}.b the agent can push the box but cannot sense it. The box neither influences the agent's outcome, nor is the agent able to perceive it. Basically, this is just like a scenario without a box.
Consequently, the empowerment map of the world is flat, i.e., all
states have the same empowerment. For empowerment applications this is
typically the least interesting case, as it provides no gradient for
action selection (see also the comment on the ``Tragedy of the Greek
Gods'' towards the end of Sec.~\ref{tragedy-of-the-greek-gods}).

Fig.~\ref{box}.d shows the empowerment map for an agent which can
perceive the box, the agent's sensor input is both its own position
and the position of the box. This different sensor configuration
changes the empowerment map of the world. Being close to the box to
affect it now allows the agent to ``reach'' more different outcomes,
because different paths that lead to the same final agent location
might affect the box differently, thereby resulting in different final
states. This results in higher empowerment closer to the box. Note
that, comparing this to the previous scenario where the box was not
visible, the agent's actions are not suddenly able to create a larger
number of resulting world states. Rather, the only change is that the
agent is now able to discriminate between different world states that
where present all along.

Figures~\ref{box}.a and ~\ref{box}.c show the empowerment map for an
non-pushable box, so when the agent moves into the box's square, its
movement fails. As opposed to the earlier cases, here we see that the
empowerment around the box is lowered, because the box is blocking the
agents way, thereby reducing the number of states that the agent can
reach with its 5-step action sequence. We also see that the
empowerment maps in Fig.~\ref{box}.a and ~\ref{box}.c are identical,
and that it does not matter if the agent can perceive the box or not.
This connects back to our earlier arguments that empowerment is about
influencing the world one can perceive. As it is not possible for the
agent to affect the box's positions, it is also not beneficial or
relevant, from an empowerment perspective, to perceive the box
position. This also relates back to earlier arguments about sensor and
motor co-evolution. Once an agent loses it ability to affect the box,
it might just as well lose it ability to sense the box.

One important insight that is demonstrated by this experiment is how
different sensor and actuator configurations can lead to significantly
different values for the state-dependent empowerment maps. Thus, which
state has the highest empowerment might depend on an agent-sensor
configuration (and not only on the world dynamics). This can be
helpful when using empowerment to define an action policy. If an agent chooses its
actions based on expected empowerment gain, then this method is a
candidate for causing an agent to change its behaviour by only
calculating empowerment for partial sensor input. For example, to
drive an agent to focus on changing its location, then selecting a
corresponding location sensor might be a good strategy.

\subsection{Horizon Extension}

\begin{figure}
\begin{center}

    \begin{minipage}[tbh]{0.24\columnwidth}
      \centering
      \fbox{\includegraphics[width=0.92\textwidth]{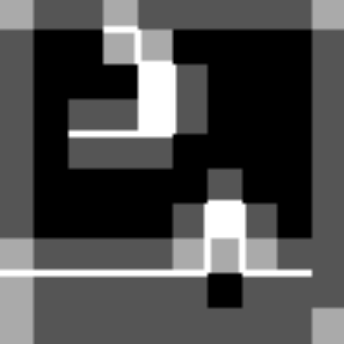}}
      \vfill
      \vspace{0.25em}
      $\mathfrak{E} \in [1; 2.32]$
      \\ 1-step Empowerment
    \end{minipage}
    \begin{minipage}[tbh]{0.24\columnwidth}
      \centering
      \fbox{\includegraphics[width=0.92\textwidth]{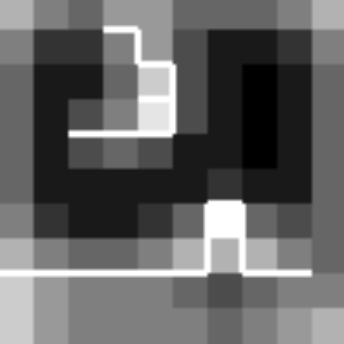}}
      \vfill
      \vspace{0.25em}
      $\mathfrak{E} \in [1.58; 3.70]$
      \\ 2-step Empowerment
    \end{minipage}
    \begin{minipage}[tbh]{0.24\columnwidth}
      \centering
      \fbox{\includegraphics[width=0.92\textwidth]{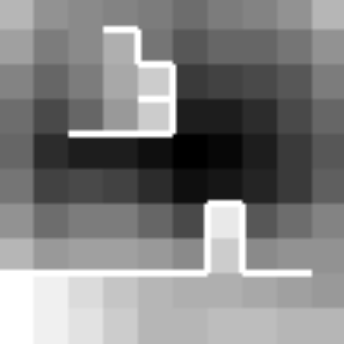}}
      \vfill
      \vspace{0.25em}
      $\mathfrak{E} \in [3.46; 5.52]$
      \\ 5-step Empowerment
    \end{minipage}
    \begin{minipage}[tbh]{0.24\columnwidth}
      \centering
      \fbox{\includegraphics[width=0.92\textwidth]{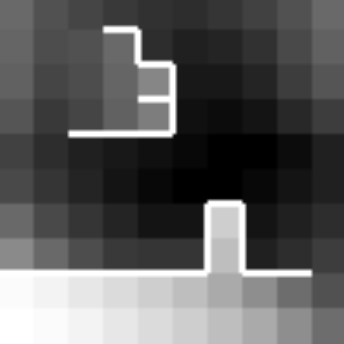}}
      \vfill
      \vspace{0.25em}
      $\mathfrak{E} \in [4.50; 6.41]$
      \\ 10-step Empowerment
    \end{minipage}
  \end{center} 
\caption{The n-step empowerment map for the same maze with different
  horizons. Figure based on \cite{klyubin2005}}
 \label{fig:mapsteps}
\end{figure}

Extending the horizon, i.e., using a larger $n$ in $n$-step
empowerment, is another way to change the actions under consideration.
Since the $n$-step action sequences can be treated just like atomic
actions, lengthening the considered sequences creates more distinct
actions to consider, which usually also have a bigger effect on the
environment. Returning to the previous maze example, Fig.~\ref{fig:mapsteps} illustrates how
the empowerment map changes for action sequences of different length.

The short-term, 1-step empowerment only takes into account its
immediate local surroundings. All that matters are if there are walls
immediately next to the agent. In general, an agent locked in a room
with walls just one step away would have the same empowerment as an
agent on an open field. Also, this map only realizes 5 different
empowerment values because the world is deterministic, and there can
be maximally 5 different outcome states.

With more steps, the empowerment map starts to reflect the immediate
surrounding of the agent and measures, as discussed by
\citet{anthony2008preferred}, how ``central'' an agent is located on
the graph of possible states. But, as discussed earlier, the world
could be shaped in a way that something just beyond the horizon of the
agent's empowerment calculation could change this completely. A
possible solution would be to further extend the horizon of the agent.
One problem, which we will address in the next section is that of
computational feasibility.

Another problem is that the agent needs the sensor capacity to
adequately reflect an increase in possible actions. Consider the
following case: computing, say, 100-step empowerment, then the agent
could reach every square from every other square, creating a flat
empowerment landscape with an empowerment of $\log(100)$ everywhere.
Since the agent itself is very (indeed maximally) powerful now, being
able to reach every state of the world, its empowerment landscape is
meaningless, as empowerment is incapable of distinguishing states via
the number of options they offer. In principle, an analogous phenomenon can
be created by massively extending the sensor capacity. Imagine an agent
would not only be able to sense it current position, but also sense
every action it has taken in the past. Now the agent could
differentiate between every possible action sequence, as every one is
reflected as a different sensor state. This again leads to a flat
empowerment landscape, with empowerment being the logarithm of all
possible actions. 

So, in short, one has to be careful when the state-space of either actions or sensors is much larger than the other. In this case it
is possible that the channel capacity becomes the maximal entropy of
the smaller variable for all possible contexts, thereby creating a flat
empowerment landscape. This phenomenon can be subsumed under the
plastic notion of the ``Tragedy of the Greek
Gods''\label{tragedy-of-the-greek-gods}: all-knowing, all-powerful
agents see no salient structure in the world and need to resort to
avatars of limited knowledge and power (in analogy to the intervention
of the Greek gods with the human fighters in the Trojan War) to attain
any structured and meaningful interaction. In short, for meaningful
interaction to emerge from a method such as an empowerment landscape,
limitations in sensing and acting need to be present. The selection of
appropriate levels of power and resolution is a current research
question.

\subsection{Impoverished Empowerment}

While seeking the right resolution for actions and sensors can be an
issue in worlds of limited complexity, a much more imminent challenge
is the fact that as the empowerment horizon grows, the number of
action sequences one needs to consider grows exponentially with the
horizon. Especially when noise is involved, this becomes quickly
infeasible.

To address this dilemma, \citet{anthony2011impoverished} suggest a
modified technique that allows for the approximation of
empowerment-like quantities for longer action sequences, arguing,
among other, that this will bring the empowerment approach in
principle closer to what is cognitively plausible.

The basic idea of the \emph{impoverished empowerment} approach is to
consider all $n$-step action sequences (as in the simple empowerment
computation), but then to select only a limited amount of
sequences from these, namely those which contribute the most to the
empowerment at this state. From the endpoint of this ``impoverished''
action sequence skeleton, this process is then repeated for another
$n$-step sequence, thereby iteratively building up longer action
sequences.

In the deterministic case, the selection is done so that the
collection of action sequences has the highest possible empowerment.
So, if several action sequences would lead to the same end state, only
one of them would be chosen.

Interestingly, a small amount of noise is useful for this process, as
it favours selecting action sequences which are further apart, because
their end states overlap less. If no noise is present, then two action
sequences which would end in neighbouring locations would be just as
valid as two that lead to completely different locations, but the
latter is more desirable as it spans a wider space of potential
behaviours.
  
\subsection{Sensor and Actuator Evolution}

Since empowerment can be influenced by the choice of sensors, it is
possible to ask what choice of sensors is maximising an agent's
empowerment. \citet{klyubin2005empowerment,klyubin2008keep} addressed
this question by using a Genetic Algorithm-based optimization for a scenario in which sensors are being evolved to maximize an agent's
empowerment. An agent is located in an infinite two-dimensional grid world. On each
turn it can take one of five different actions which are to move
in one of four directions, or to do nothing. Each location now has a
value representing the concentration of a marker substance which is
inversely proportional to the distance of the current location to the
center at 
location $(0,0)$.

In this scenario, the agents sensors can change, both in positioning
and number. A sensor configuration is defined by where each of the $n$
sensors of the agent is located relative to the agent. The sensor
value has $n$ states, and represents which of the $n$ sensors detects
the highest concentration value of the marker.

\begin{figure}
\centering
\includegraphics[scale=1]{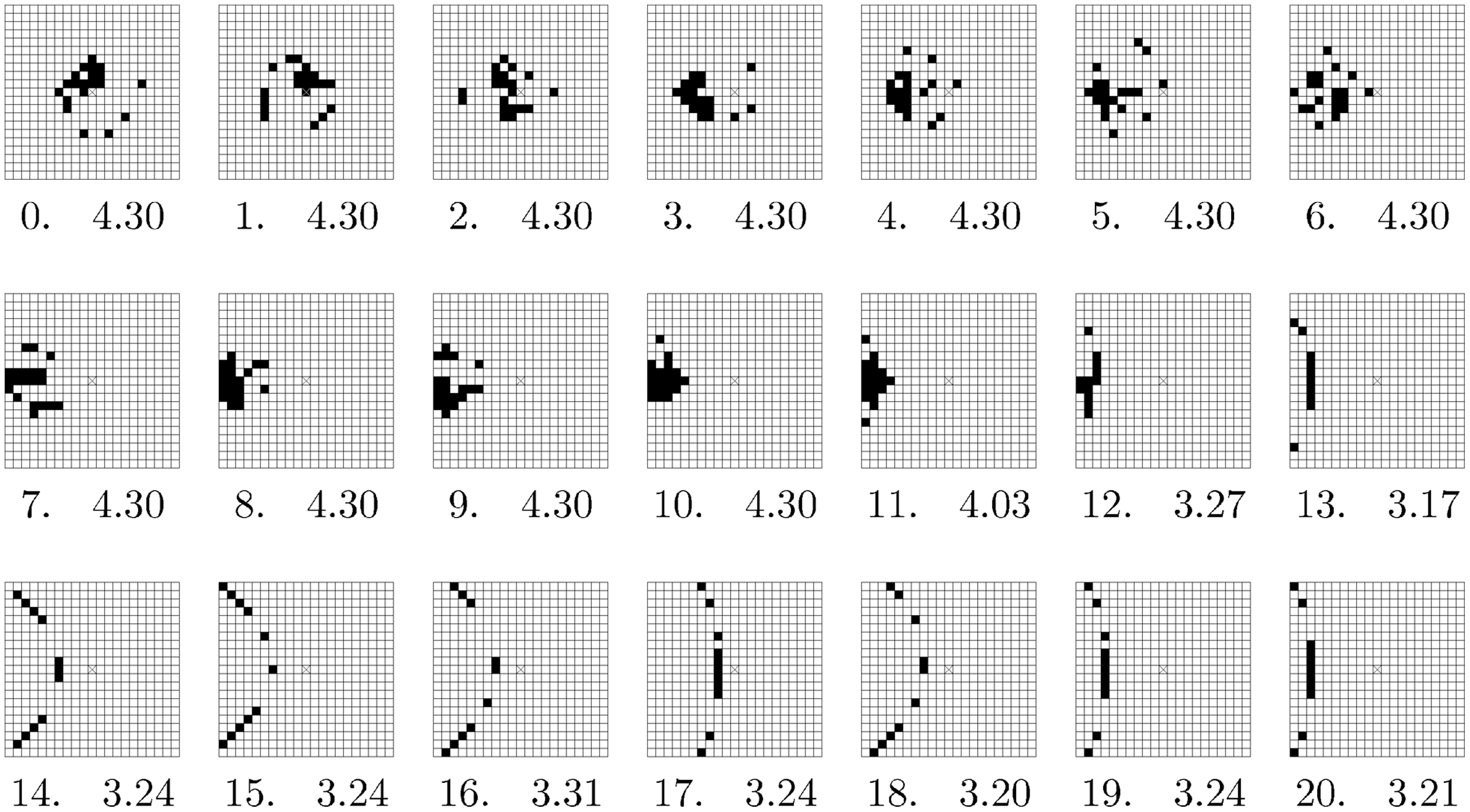}
\caption{The Figures show what sensor configurations empowerment evolves for different starting positions. The first number indicates how many spaces east of the center the agent starts, and the second number is the resulting empowerment value of the sensor configuration.    Figure taken from \cite{klyubin2008keep}}
\label{SensorEvolution}
\end{figure}

\citet{klyubin2005empowerment,klyubin2008keep} then evolved the agents
sensor configuration to maximise empowerment for different starting
locations with respect to the centre. So, for example, they evolved
the sensor configurations to achieve the highest empowerment when the
agent starts its movement at location $(0,0)$. To avoid degeneracy, a
slight cost factor for the number of sensors was added. In this way
the adaptation has to evaluate if the added cost of further sensors
are worth the increase in empowerment. The resulting sensor
configurations for a 4-step empowerment calculation can be seen in
Figure \ref{SensorEvolution}.
 
The result was unsurprisingly that different starting positions would
lead to different sensor layouts. More interestingly, they
realized that the space of possible solutions can be more constrained
in some places, so there is only one good solution, while other locations offer
several different, nearly equally empowered solutions. More
importantly is the observation that empowerment agnostically
selects modalities which are most appropriate for the various starting
locations. Consider, for instance, Fig.~\ref{SensorEvolution} which
shows how the sensors are placed relative to the agent as the agent
moves increasingly away from the center of the world, and to the right
of it. The first images show the sensor placement when the agent is at
the center of the world. The sensors are placed with more-or-less
precision around the center, and there is some indifference as to
their exact placement. 

In the second row, when the agent has been
moved seven and more fields to the right of the centre, a more
prominent ``blob'' is placed at around the location of the centre (the
diagram shows the relative placement of the sensors with respect to
the agent, so a blob of black dots is covering roughly the location at
which the centre of the world will be with respect to the agent.

Finally, as the agent moves further to the right (end of second and last row in  Fig.~\ref{SensorEvolution}), a striking effect takes place: the blob sensor,
which roughly determines a two-dimensional location of the centre,
collapses into a ``heading'' sensor which is no longer a
two-dimensional blob, but rather has 1-dimensional character. This
demonstrates that empowerment is able to switch to different
modalities (or, in this case, from a 2-dimensional to a 1-dimensional
sensor). Because of its information-theoretic nature, empowerment is
not explicitly using any assumptions about modality or dimensionality
of sensors. The resulting morphologies are purely a result of the
selection pressure via empowerment in interaction with the dynamics
and structure of the world under consideration.

Another result of the evolutionary scenario involved the evolution of
actuators. Without repeating the full details that can be found in
\cite{klyubin2008keep}, we would like to mention one important result,
namely that the placement of actuators via empowerment-driven
evolution, unlike the sensors, was extremely unspecific. Many
configurations led to maximum empowerment solutions. The authors suggest that this results as a consequence of the agent being unable to choose what form the 'information' takes, that it has to extract from the environment. Hence, the sensors have to adapt to the information structure available in the environment, leaving the agent free to choose its actions. Therefore many different actuator settings can be
used as the agent can utilize each of them to full effect by
generation of suitable action sequences. This is an indicator that an
agent's action choices should be a more valuable and ``concentrated''
source of information than the information extracted from the
environment, as every action choice is significant, while sensoric
information needs to be ``scooped'' in on a wide front to capture some
relevant features. This insight has been taken onboard in later work
in form of the the concept of \emph{digested information} \cite{salge2011digested} where agents observe other agents
because their actions are more informationally dense than other
aspects of the environment. The core idea of \emph{digested} information is that relevant information (as defined in \cite{polani2006relevant}) is often spread out in the environment, but since an agent needs to act upon the information it obtains, the same information is also present in the agent's actions. Because the agent's action state-space is usually much smaller than the state-space of the environment, the agent ``concentrates'' the relevant information in it actions. From the perspective of another, similar agent this basically means that the agent digests the relevant information and then provides it in a more compact format. It should be noted that all structure in the above example emerges purely from informational considerations; no
other cost structure (such as e.g.\ energy costs) have been taken into
account to shape the resulting features.

\subsection{Multi-Agent Empowerment}

If two or more agents share an environment, so that their actions all
influence the state of the world, then their empowerment becomes
intertwined.
\citet{capdepuy2010,capdepuy2007maximization,capdepuy2012perception}
investigate this phenomenon in detail. Here, due to lack of space, we
will limit ourselves to briefly outline his results.

If both agents selfishly optimize their empowerment, then the outcome
depends heavily on the scenario. A fully formal categorization is
still outstanding, but the qualitative phenomenon can be described in
terms similar to different game solution types in game theory. One
finds situations that are analogous to zero-sum games where the
empowerment of one agent can only be raised to the detriment of the
other. In other situations, selfish empowerment maximisation leads to
overall high empowerment, and, finally, there are scenarios where
agent's strategies converge onto the analoga of intricate equilibria
reminiscent of the Nash equilibria in games.
  
\begin{figure}
  \centering
  \includegraphics[scale=0.4]{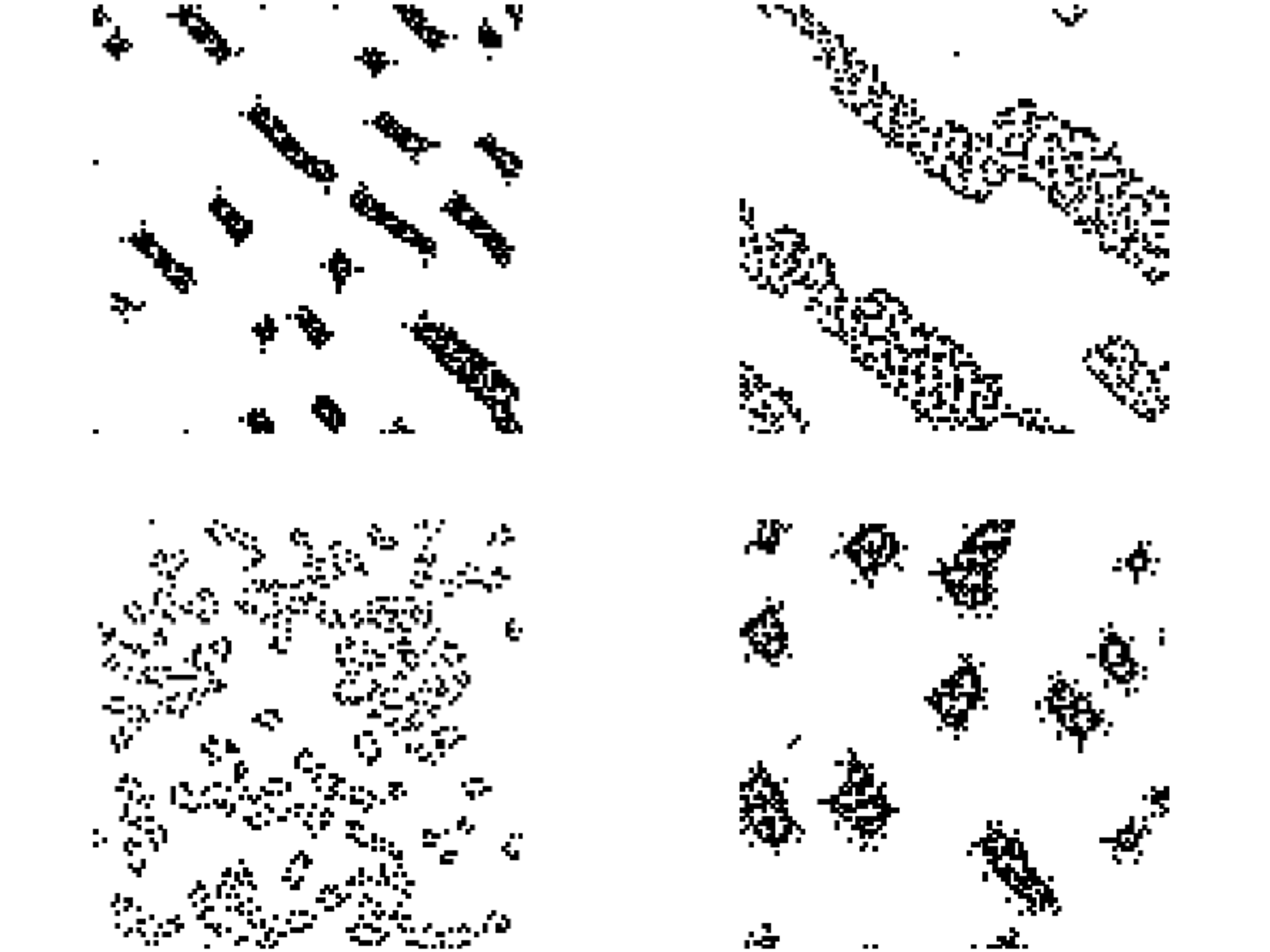}
  \caption{Structures resulting from agent behaviour that was evolved
    to maximise the agents' individual empowerment. Each black dot in
    the figure represents an agent in one of the
    empowerment-maximizing scenarios. Agents are equipped with
    directional density sensors, measuring the number of other agents
    present in that particular direction. Creating structures becomes
    beneficial for the agents, as it gives features to the environment
    that allow different resulting sensor inputs. The different structures are high empowered solutions of the artificial evolution.  Figure taken from
    \cite{capdepuy2010}}.
  \label{StructureEvolution} 
\end{figure}

An interesting aspect in relation to biology is Capdepuy's work on the
emergence of structure from selfish empowerment
maximisation \cite{capdepuy2007maximization}. The model consists of a
two-dimensional grid world where agents are equipped with sensors that
measure the density of other agents in the directions around them. In
this case, there is a tension between achieving proximity to other
agents (to attain any variation in sensor input, as empty space does
not provide any) and being sufficiently distant (as to attain
sufficient freedom for action and not to be stuck without ability to
move); this tension, in turn, provides an incentive to produce
nontrivial dynamical structures. Some examples of agent populations
evolved for greedy empowerment maximization and some of the better
empowered structures resulting from this process can be seen in
Fig.~\ref{StructureEvolution} \citet{capdepuy2007maximization}.

\section{Continuous Empowerment}
\label{sec:cont-empow}
The empowerment computations that we considered earlier were all
operating in discrete spaces. But if we want to apply empowerment to the real world we need to consider that many problems, especially those related to motion or motor control, are continuous in nature. We could apply naive discretizations with finer and finer resolutions, but this will quickly lead to large state and actions spaces, with a forbidding number of options where direct computation of empowerment become very computationally expensive
\cite{klyubin2008keep}; therefore, different approaches need to be
taken to deal with continuous dynamics effectively. 

In this section, we will take a closer look at empowerment for
continuous actuator and sensor variables. Compared to the discrete
case, while channel capacity is still well defined for continuous
input/output spaces, there are some important conceptual differences
to be considered as compared to the discrete case.

One problem, as we shall illustrate, is that the continuous channel
capacity could --- in theory --- be infinite. The reason for this is
as follows: if there is no noise, and arbitrary continuous actions can
be selected, these actions now allow to inject continuous, i.e.\
real-valued quantities (or vectors) into the world state. Reading in
their (again) noiseless effect through real-valued sensors means that
the full precision of a real number can be used in such a case. As
arbitrary amounts of information can be stored in an infinite
precision --- noiseless --- real number, this implies (in
nondegenerate cases) an infinite channel capacity. Of course, such a
situation is not realistic; in particular, relevant real-world systems
always have noise and therefore the channel capacity will be limited.

However, when modeling a deterministic system with floating-point
precision in simulation, there is no natural noise level. In a
nondegenerate system, empowerment can be made as large as the
logarithm of the number of actions (action sequences) available. This is,
of course, meaningless. To be meaningful, one needs to endow the
system with additional assumptions (such as an appropriate noise
level) 
which are not required in the deterministic case.


But the main problem in the continuous case is that there is at the
time of this review no known analytic solution to determine the
channel capacity for a general continuous channel. To address this
problem, a number of methods to approximate continuous channel
capacity have been introduced. We will discuss them and how they can
be used to compute empowerment.

We will briefly discuss naive binning, then the Monte Carlo
Integration method developed by \cite{jung2011empowerment}, and then
focus mostly on the quasi-linear Gaussian approximation, which is fast
to compute.

\subsection{Continuous Information Theory}

The analogy to discrete entropy is rigorously defined for continuous
random variables as \emph{differential entropy}
\begin{equation}
h(X) = - \int_{\mathcal{X}}p(x)\log(p(x)) \ \mathrm{d}x\;,
\end{equation}
where $p(x)$ now denotes not the probability, but the probability
density function of $X$, defined over a support set of $\mathcal{X}
\subset \mathbb{R}$. Similarly, the \emph{conditional differential
  entropy} is defined as
\begin{equation}
\label{eq:2}
h(X|Y) = - \int_{\mathcal{Y}}p(y)
\int_{\mathcal{X}} p(x|y)\log(p(x|y)) \ \mathrm{d}x \mathrm{d}y\;.
\end{equation}
The differential entropies cannot be directly interpreted in the same
way as discrete entropies: they can become infinite or even negative.
However, without delving too much into their individual interpretation,
we will just state here that the difference of two differential
entropy terms again can be interpreted as a proper mutual
information: 
$I(X;Y) := h(X) - h(X|Y)$, which shares essentially all characteristics of
the discrete mutual information\footnote{One exception is that the
  continuous version of mutual information can become infinite in the
  continuum. This, however, is perfectly consistent with the ability
  to store infinite amount of information in continuous variables and
  does not change anything substantial in the interpretation.%
}. Thus, consequently, the channel capacity is again defined by
maximising the mutual information for the input probability density
function
\begin{equation}
\mathfrak{E} = C(A\rightarrow S) = \max_{p(a)} I(A;S).
\end{equation}
We will still be dealing with discrete time steps. Just like in the
discrete case, we will use the notation $A_t$ and $S_t$ not just for
single, but also for compound random variables. So, for each time $t$,
both variables $A_t$ and $S_t$ can consist of vectors of multiple
random variables. The variables $A$ and (where relevant) $S$ itself
are then again a selection of actuator and sensor variables at
different times $t$, so for example, the actuator input for $n$-step
empowerment might be written compactly as $A=(A_{t},...,A_{t+n-1})$.

\subsection{Infinite Channel Capacity}

As mentioned above, in contrast to the discrete case, the continuous
channel capacity can be infinite for some $p(s|a)$. Formally, this
results from the fact that differential entropy \textit{can} become
negative. For instance, it becomes negative infinity for a Dirac
$\delta_x(.)$ ``distribution''. The Dirac ``distribution'' is a
probability measure concentrated on a single point: it can be
mathematically defined in a precise fashion, but for the following
discussion, the intuition is sufficient that $\delta_x(.)$ is
normalized (the integral over this ``distribution'' is 1), and is 0
everywhere with exception of the one point $x$ at which it is
concentrated, where it assumes an infinite value.

To illustrate, imagine that the channel $p(s|a)$ exactly reproduces
the real-valued input value of $a \in \mathbb{R}$, i.e.\ that it
implements $s=a$, i.e.\ $p(s|a) \equiv \delta_a(s)$. Every input $a$
precisely determines the output $s$, so $h(S|a)=-\infty$. This remains
negative infinity when we integrate over all possible inputs, so
$h(S|A)=-\infty$. If we now choose for
$p(a)$ the uniform input distribution between $0$ and $1$, which has a
differential entropy of $0$, we then get the following mutual
information\footnote{Strictly spoken, we should denote this quantity
  as \emph{differential} mutual information, but unlike the
  differential entropy, this term retains the same interpretation in
  the continuous as in the discrete case, and therefore we will not
  especially qualify it by terming it ``differential''.}
\begin{equation}
I(A;S) = h(S)- h(S|A) = h(A) - (-\infty)=\infty\;.
\end{equation}     
It holds $H(S)=H(A)$, because the channel just copies the input
distribution to the output. Since this is the largest possible value,
this is also the channel capacity.

%


\subsection{Continuous Empowerment Approximation}

While channel capacity is well defined for any relationship between
$S$ and $A$, it can only be computed for a subset of all possible
scenarios. We will here approximate the model of the world with one
for which empowerment can be computed. The following section discusses
different approaches for doing so.

\subsection{Binning}

The most straightforward and naive approximation for continuous
empowerment is to discretize all involved continuous variables and
then compute the channel as described in the discrete empowerment
section. 



However, there are different ways to bin real-valued numbers and, as
\citet{olsson2005sensor} demonstrated, they clearly affect the
resulting informational values. \textit{Uniform binning} considers the
support of a real-valued random variable (i.e.\ the set of values of
$x$ for which $p(x) > 0$), splits it into equally sized intervals and
assigns to each real number the bin it falls into. Of course, this
does not necessarily result in the same number of events in each bin
and, furthermore, many bins can be left empty or with very few events
while others contain many events. This unevenness can mean that
significant ``information'' (in the colloquial sense) in the data is
being discarded. The response is to choose the binning in a not
necessarily equally spaced way, that ensures that all bins are used,
and that the events are well distributed. This is achieved by
\textit{Max-Entropy binning} where one adaptively resizes the bins so
the resulting distribution has the highest entropy, which usually
results in bins containing the approximately same number of
events \citet{olsson2005sensor}.

There are two caveats for this case: If adaptive binning is chosen,
one needs to take care that the informational values of different
measurements are comparable, and that the binning is the same
throughout the same context of use. Therefore, it is important to
choose the binning in advance, say, adapted only to the overall,
context-free channel, and not adapt to each state-dependent channel
separately. The second caveat is that, while adaptive binning
distributes the events more-or-less evenly over the bins, this can
thin out the sampling very considerably and cause the bins to be
almost empty or containing very few elements each. This can induce the
appearance of nonzero mutual information which, however, is spurious.
In this case, it is better to choose a binning that is wide enough to
ensure a sufficient number of events per bin. 
Both approaches require the availability of actual samples, so if the
channel in question is only specified as a continuous conditional
probability, it is necessary to generate random samples based on
$p(s|a)$.

A final note on information estimation: much more robust approaches
for mutual information estimation are known, such as the
Kraskov-St\"ogbauer-Grassberger (KSG) estimator
\cite{PhysRevE.69.066138}. Unfortunately, this method is not suitable
for use with empowerment, as it requires the full joint distribution
of the variables to be given in advance. When computing empowerment,
however, one iteratively selects an input distribution, computes a
joint distribution and then applies the information estimator. This
means that if one uses the KSG-estimator, it affects the joint
distributions and hence its own estimates of mutual information at
later iterations of the process, and thus the conditions for correct
operation of KSG cease to  hold\footnote{The authors thank Tobias
  Jung for this information (private communication).}.

%

\subsection{Evaluation of Binning}

One problem with this approach is that it can introduce binning
artefacts. Consider the following example: imagine one bins by proper
rounding to integers. In this case, outcomes such as, say, 0.6 and 1.4
become the same state, while 1.4 and 1.6 are considered different. If
now an agent which moves along the real valued line by an amount of
0.2 at each time step, this binning would make the agent appear to be
more empowered at 1.5 then it would be at 1.0, because it could move
to two different resulting states from 1.5. If the binning would
reflect true sensoric resolution of the agent, this would conform with
the empowerment model of being able to resolve the corresponding
states; however, in our example, we did not imply anything like that
--- the underlying continuous structure is completely uniform, and we
did not introduce any special sensoric structure. Thus, the difference
in empowerment is a pure artefact introduced by the binning itself.

Another problem that emerges with the use of a binning approach is the
right choice of granularity. If too few bins are chosen, then, while
one has a good number of samples in the bins, interesting structural
effect and correlations are lost. If too many bins are chosen, then
many (or all bins) contain very few samples, perhaps as few as only
one or even none. Such a sparse sampling can significantly
overestimate the mutual information of the involved variables. Another
problem, specifically in conjunction with empowerment, is that such a
sparse sampling is often likely to cause one action to produce exactly
one distinguishable sensoric outcome. This means that empowerment
reaches its maximum $\log |\mathcal{A}|$ for every context $r$
depriving it of any meaning.
However, if the resolution is high enough and sufficiently many
samples are collected, binning can produce a quickly implemented (but
typically slow to compute) approximation for empowerment. Examples of its
application to the simple pendulum can be seen in \cite{klyubin2008keep}.

\subsection{Jung's Monte Carlo Integration}
\label{JMCI}

Another approximation to compute empowerment which can still deal with
any kind of $p(s|a)$ is Monte Carlo Integration
\cite{jung2011empowerment}. It is computed by sampling the outcomes of
applying a representative set of available action sequences.

Assume that you have a model, so for a state $r$ you can take actions
$a_v$, with $v = 1,...,n$, and draw $N_{MC}$ samples, which will
result in sensor states $s_{v,j}$, with $j=1,...,N_{MC}$. This method
then approximates the term $d_{v,k}$ from Eq.~(\ref{dvk}) in the BA
by
\begin{equation}
d_{v,k} \approx \frac{1}{N_{MC}} \sum^{N_{MC}}_{j=1} \log \left[  \frac{p(s_{v,j}|r,\vec{a}_{v})}{\sum^{n}_{i=1} p(s_{v,j} | r, \vec{a}_{i}) p^{i}_{k}} \right].
\label{MCIntegration}
\end{equation}

To compute this the model needs to provide a way to compute how
probable it is that the outcome of one action was produced by another.
The necessary noise in the model basically introduces a ``distance
measure'' that indicates how hard it is to distinguish two different
actions.

One simple model is to assume that $p(s|r,\vec{a}_{v})$ is a
multivariate Gaussian (dependent on the current state of the world $r$), or can be reasonably well-approximated by it,
i.e.,
\begin{equation}
\label{GaussAssumption}
s| r, \vec{a_{v}} \sim \mathcal{N}(\mathbf{\mu}_{v},\mathbf{\Sigma}_{v})
\end{equation}
where $\mathbf{\mu}_{v} = (\mu_{v,1},...,\mu_{v,n})^{T}$ is the mean
of the Gaussian and the covariance matrix is given by
$\mathbf{\Sigma}_{v} = diag (\sigma^{2}_{v,1},...,\sigma^{2}_{v,n})$.
The mean and covariance will depend upon the action $\vec{a}_{v}$ and
the state $r$. Samples from the distribution will be denoted
$\Tilde s_v$ and can be generated using standard algorithms.

The following algorithm summarises how to approximate the empowerment
$\mathfrak{E}(r)$ given a state $r \in \mathcal{R}$ and transition
model $p(s|r,\vec{a}_{v})$:

\begin{enumerate}
	\item{\textbf{Input:}}
	\begin{enumerate}
		\item{Specify state $r$ whose empowerment is to be calculated.}
		\item{For every action $a_v$ with $v = 1,...,n$, define a (Gaussian) state transition model $p(s| r, \vec{a}_{v})$, which is fully specified by its mean $\mathbf{\mu}_{v}$ and covariance $\Sigma_{v}$.}
	\end{enumerate}
	\item{\textbf{Initialise:}}
	\begin{enumerate}
		\item{$p_{0}(\vec{a}_{v}) := 1/n$ for $v=1,...,n$.}
		\item{Draw $N_{MC}$ samples $\tilde{s}_{v,i}$ each, according to distribution density $p(s|r,\vec{a}_{v}) = \mathcal{N}(\mathbf{\mu}_{v},\mathbf{\Sigma}_{v})$ for $v = 1,...,n$.}
		\item{Evaluate $p(\tilde{s}_{v,i}|r,\vec{a}_{\mu})$ for all $v = 1,...n$; $\mu = 1,...n$; and sample $i = 1,...,N_{MC}$.}
	\end{enumerate}
	\item{\textbf{Iterate} the following variables for $k = 1, 2, ...$ until $|\mathfrak{E}_{k} - \mathfrak{E}_{k-1}| < \epsilon$ or the maximum number of iterations is reached:}
	\begin{enumerate}
		\item{$z_{k} := 0$, $\mathfrak{E}_{k-1} := 0$}
		\item{For $v=1,...,n$
		\begin{enumerate}
			\item{$d_{v,k} :=\displaystyle \frac{1}{N_{MC}} \sum^{N_{MC}}_{j=1} \log \left[\displaystyle \frac{p(\tilde{s}_{v,j}|r,\vec{a}_{v})}{\sum^{n}_{i=1} p(\tilde{s}_{v,j} | r, \vec{a}_{i}) p^{i}_{k}} \right]$}
			\item{$\mathfrak{E}_{k} := \mathfrak{E}_{k-1} + p_{k-1}(\vec{a}_{v}) \cdot d_{v,k-1}$}
			\item{$p_{k} := p_{k-1}(\vec{a}_{v}) \cdot  \exp (d_{v,k-1})$}
			\item{$z_{k} := z_{k} + p_{k}(\vec{a}_{v})$}
		\end{enumerate}}
		\item{For $v=1,...n$
		\begin{enumerate}
			\item{$p_{k}(\vec{a}_{v}) := p_{k}(\vec{a}_{v}) \cdot z^{-1}_{k}$}
		\end{enumerate}}
	\end{enumerate}
	\item{\textbf{Output:}
	\begin{enumerate}
		\item{Empowerment $\mathfrak{E}(r) \approx \mathfrak{E}_{k-1}$ (estimated).}
		\item{Distribution $p(\vec{a})$ achieving the maximum mutual information.}
	\end{enumerate}}
\end{enumerate}

\subsection{Evaluation of Monte Carlo Integration}

Monte Carlo Integration can still deal with the same generic distributions $p(s|a)$ as the binning approach, and it removes the artefacts caused by the arbitrary boundaries of the bins. On the downside, it  requires a model with a noise assumption. In the solution suggested by \citet{jung2011empowerment} this lead to the assumption of Gaussian Noise.

The other problem is computability. For good approximations the number of selected representative action sequences should be high, but this also leads to a quick growth of computation time. The several applications showcased in \cite{jung2011empowerment} all had to be computed off-line, which makes them infeasible for robotic applications.   

\subsection{Quasi-Linear Gaussian Approximation}


In the previous section we saw that Jung's Monte Carlo Integration
method could deal with the rather general case where the relationship
between actuators and sensor can be characterized by $s = f(r,a) + Z$,
where $f$ is a deterministic mapping, and $Z$ is some form of added
noise. The noise is necessary to limit the channel capacity, and an
integral part of the Monte Carlo Integration in
Eq.~\ref{MCIntegration}. While the noise can have different
distributions, Jung's example assumed it to be Gaussian.

We will now outline how the assumption of Gaussian noise, together
with an assumption regarding the nature of $f$, will allow us to
accelerate the empowerment
approximation. 
Consider now actuation-sensing mappings of the form $s = f(r,a) +
\mathcal{N}(0,N_r)$, i.e.\ which can be described by a deterministic
mapping $f$ on which Gaussian noise (which may depend on $r$) is
superimposed\footnote{We will treat this as centred noise, with a mean
  of 0, but this is not necessary, as any non-zero mean would just
  shift the resulting distribution, which would leave the differential
  entropies and mutual information unaffected.}.

In principle, if the actions $A$ were distributed in an arbitrarily
small neighbourhood around 0, one would need $f$ to be differentiable
in $a$ with the derivative $D_af$ depending continuously on $r$. In
practice, that neighbourhood will not be arbitrarily small, so the
mapping from $a$ to $s$ needs to be ``sufficiently well'' approximated
at all states $r$ by an affine (or shifted linear) function in
$f_r(a)$ for the allowed distributions of actions $p(a)$. To limit the
channel capacity there has to be some constraint on the possible
action distributions, and the linear approximation has to be
sufficiently good for the actions that $A$ can actually
attain.\footnote{We will not make this notion more precise or derive
  any error bounds at this point; we just informally assume that the
  Gaussian action distribution $A$ is concentrated well enough for
  $f_r$ to appear linear in $a$.}



In other words, assuming the channel can be adequately approximated by a
linear transformation applied to $A$ with added Gaussian noise, then
it is possible to speed up the empowerment calculation significantly
by reducing the general problem of continuous channel capacity to
parallel Gaussian channels which can be solved with well-established
algorithms. This provides us with the \emph{quasi-linear Gaussian
  approximation} for empowerment which will now be presented in
detail.

Let $S$ be a multi-dimensional, continuous random variable defined
over the vector space $\mathbb{R}^n$. Let $A$ be a multidimensional
random variable defined over $\mathbb{R}^m$. As in the discrete, $A$
is the action variable, and $S$ the perception variable. According to
the quasi-linear Gaussian approximation assumption, we assume that
there is a linear transformation $T: \mathbb{R}^m \rightarrow
\mathbb{R}^n$ that allows us to express the relation between these
variables via 
\begin{equation}
S = TA + Z.
\label{eq:mapping}
\end{equation}
$Z$ is a suitable multi-dimensional, Gaussian variable defined over
$\mathbb{R}^n$, modelling the combined acting/sensing noise in the
system and is assumed to be independent of $A$ and $S$.


Consider first the simpler white noise case. Here we assume that the
noise in each dimension $q \leq n$ of $Z$ is independent of the noise
in all other dimensions, and has a normal distribution with $Z_q \sim
\mathcal{N}(0,N_q)$ for each dimension (where $N_q$ depends on the
dimension). This particular form of noise can be interpreted as having
$n$ sensoric channels where each channel $q$ is subject to a source of
independent Gaussian noise.


We now further introduce a limit to the \emph{power} $P$ available to
the actions $A$, i.e.\ we are going to consider only action
distributions $A$ with $E(A^2) \leq P$. The reason for that is that
without this constraint, the amplitude of $A$ could be made
arbitrarily large and this again would render all outcomes
distinguishable and thus empowerment infinite\footnote{This specific
  power limit also implies that the optimal input distributions for
  the channel capacity results is Gaussian \cite{Cover1991}.}. The
actual mean of the distributions is irrelevant for our purpose, as a
constant shift does not affect the differential entropies. However, we
need to ensure that the actuation range considered does not extend the
size for which our linearity assumption holds.

It is plausible to consider this limitation as a physical power
constraint\footnote{This point is subtle: throughout the text, we had
  made a point that empowerment is determined by the structure of the
  actuation-perception loop, but otherwise purely informational. In
  particular, we did not include any further assumptions about the
  physics of the system. In the quasi-linear Gaussian case, the choice
  of a ``physics-like'' quadratic form of power limitation is only
  owed to the fact that it makes the problem tractable. Other
  constraints are likely to be more appropriate for a realistic
  robotic actuator model, but need to be addressed in future work.}.
Under these constraints, the quantity of interest now becomes
\begin{equation}
\mathfrak{E} = \max_{p(a) : E(A^2)\leq P} I(S;A)
\label{eg:channelCapacity}
\end{equation}
and the maximum being attained for normally distributed $A$ (thus we
only need to consider Gaussian distributions for $A$ in the first
place).

\subsection{MIMO channel capacity}
\label{sec:Mimo}

Now, assume for a moment that, in addition to our assumption of
independent noise, the variance of the noise $Z$ in each dimension has
the same value, namely 1, then the problem becomes equivalent to
computing the channel capacity for a linear,
Multiple-Input/Multiple-Output channel with additive and isotropic
Gaussian noise. Though the methods to compute this quantity are well
established in the literature, for reasons of self-containedness, we reiterate them here. 

The MIMO problem can be solved by standard methods
\cite{telatar1999capacity}, namely by applying a Singular Value
Decomposition (SVD) to the transformation matrix $T$, which decomposes
$T$ as
\begin{equation}
T = U \Sigma V^{T}
\end{equation}
where $U$ and $V$ are unitary matrices and $\Sigma$ is a diagonal
matrix with non-negative real values on the diagonal. This allow us to
transform Eq.~(\ref{eq:mapping}) to
\begin{equation}
U^{T} S = \Sigma V^{T}A + U^{T}Z.
\label{SVDtransformed}
\end{equation}
It can be shown that each dimension of the resulting vectorial
variables $U^{T} S$, $\Sigma V^{T}A$ and $U^{T}Z$ can be treated as an
independent channel (see \cite{telatar1999capacity}), and thus
reducing the computation of the overall channel capacity to computing
the channel capacity for linear, parallel channels with added Gaussian
noise, as in \cite{Cover1991},
\begin{equation}
C =\max_{P_i} \sum_i {\frac{1}{2}\log\left( 1 +
    \frac{\sigma_iP_i}{E\left[ (\mbox{$U$}^{T} Z)^2_i\right]}\right)} =\max_{P_i}\sum_i {\frac{1}{2}\log(1 + {\sigma_iP_i}})
\label{capacity}
\end{equation}
where $\sigma_i$ are the singular
values of $\Sigma$, and $P_i$ is the average power used in the $i$-th
channel, following the constraint that
\begin{equation}
\sum_i{P_i} \leq P.
\end{equation}
The simplification in the last step of Eq.~(\ref{capacity}) is based on the assumption of isotropic noise. Because the expected value for the noise is 1.0 and the unitary matrix applied to $Z$ does not scale,
but only rotates $Z$, the noise retains its original value of 1.0.

We remind that the channel capacity achieving distribution for a
simple linear channel with added Gaussian Noise is Gaussian
\cite{Cover1991}. In particular, the optimal input distribution for
each subchannel is a Gaussian with a variance of $P_i$. The optimal
power distribution which maximizes Eq.~(\ref{capacity}) can then be
found with the \emph{water-filling algorithm} \cite{Cover1991}. The
basic idea is to first assign power to the channel with the lowest
amount of noise. This has an effect that could be described as one of
``diminishing returns'': once a certain power level is reached, where
adding more power to that channel has the same return as adding to the
next best channel,
additional power is now allocated to the \emph{two} best channels.
This is iterated to the next critical level and so on, until all power
is allocated. Depending on the available total power, not all channels
necessarily get power assigned to them.

We can also see, directly from the formula in Eq. (\ref{capacity}),
that since we divide by the variance of the noise $Z$, this value needs to be larger than zero. For
vanishing noise, the channel capacity becomes infinite. Only the
presence of noise induces an ``overlap'' of outcome states that allows
one to obtain meaningful empowerment values. However, this is not a
significant limitation in practice, as virtually all applications need
to take into account actuator, system and/or sensor noise.

\subsection{Coloured Noise}

In a more general model, the Gaussian noise added to the multi-inputs,
multi-output channel might also be coloured, meaning that the noise
distributions in the different sensor dimensions are not
independent. Let us assume that the noise is given by $Z \sim
\mathcal{N}(0,K_s)$, where $K_s$ is the covariance matrix of the
noise. As above, we assume that the distribution has a mean of zero,
which is without loss of generality since translations are information
invariant. The relationship between $S$ and $A$ is again expressed as
 \begin{equation}
S = T'A + Z'.
\end{equation}
Conveniently, this can also be reduced to a channel with i.i.d.\
noise. For this, note that rotation, translation and scaling operators
do not affect the mutual information $I(S;A)$. We start by expressing
$Z'$ as
\begin{equation}
Z' = U\sqrt{\Sigma} ZV^T,
\end{equation} 
where $Z \sim \mathcal{N}(0,I)$ is isotropic noise with a variance of
1, and $U\Sigma V^T = K_s$ is the SVD of $K_s$. $U$ and $T$ are
orthogonal matrices, and $\Sigma$ contains the singular values. Note
that all singular values have to be strictly larger than zero,
otherwise there would be a channel in the system without noise, which
would allow the empowerment maximizer to inject all power into the
zero-noise component of the channel and to achieve infinite channel
capacity. $\sqrt{\Sigma}$ is a matrix that contains the square roots
of the singular values, which should scale the variance of the
isotropic noise to the singular values. The orthogonal matrices then
rotate the distributions, so that they resemble $Z'$.

If we consider $\sqrt{\Sigma}^{-1}$, a diagonal matrix whose entries
are the inverse of the square root of the singular values in $\Sigma$,
this allows us to reformulate:
\begin{eqnarray}
S &=& TA + U\sqrt{\Sigma} ZV^T \\
U^TS V &=& U^T T A V + \sqrt{\Sigma} Z \\
\sqrt{\Sigma}^{-1}U^TS V &=& \sqrt{\Sigma}^{-1}U^TT A V + Z \\
\sqrt{\Sigma}^{-1}U^TS &=& \sqrt{\Sigma}^{-1}U^TT A + ZV^T \\
\sqrt{\Sigma}^{-1}U^TS &=&\sqrt{\Sigma}^{-1}U^TT A + Z 
\end{eqnarray}
The last step follows from the fact that the rotation of isotropic
Gaussian noise remains isotropic Gaussian noise. This reduces the
whole problem to a MIMO channel with isotropic noise and with the same
channel capacity. We simply redefine the transformation matrix $T$ as
\begin{equation}
T = \sqrt{\Sigma}^{-1}U^T T',
\end{equation}
and solve the channel capacity for $S = TA + Z$, as outlined in section \ref{sec:Mimo}.

\subsection{Evaluation of QLG Empowerment}

The advantage of the quasi-linear Gaussian approximation is that it is
quick to compute, the computational bottleneck being the calculation
of a singular value decomposition that has the same dimensions as the
sensors and actuators.

The drawbacks are its introduction of several assumptions. Like Jung's
integration, the approximation forces us to assume Gaussian noise.
However, a more aggressive assumption than Jung's approximation is
that the QLG approximation also needs a locally linear model. So it is
not possible to represent locally non-linear relationships between the
actions and sensors. In particular the abrupt emergence of novel
degrees of freedom which the empowerment formalism is so apt at
discovering (see above, e.g. box pushing in Sec.~\ref{boxpushing}) becomes softened by
the Gaussian bell of the agent's actuations.

Finally, the quasi-linear Gaussian approximation also introduces a new
free parameter, the power-constraint $P$ which will be discussed in a
later example. A more detailed examination of QLG empowerment can be
found in \cite{Salge2012}.

\section{Continuous Examples}

We are aware of currently only two publications dealing with
continuous empowerment. The first, by \citet{jung2011empowerment},
provides a good technical tutorial, and introduces the Monte Carlo
Integration technique. Furthermore, it
demonstrates that those states generally chosen as goals have high
state-dependent empowerment, and that an empowerment-driven controller
will tend to drive the system into them, even when initialized from a
far away starting point. So, for example, the simple pendulum swings
up, and stabilizes in the upright position, even when multiple
swing-up phases are required; unlike traditional Reinforcement
Learning, there is no value function that needs to be learnt over the
whole phase space, but only the transition dynamics, and that needs
only to be determined around the actual path taken. In principle, the
algorithm does not need to visit any states but those in the
neighbourhood of the path taken by the empowerment-driven controller.
The empowerment-driven control method can be applied also to other,
quite more intricate models, such as bicycle riding or the acrobot
scenario (double-pendulum hanging from the top joint and driven by a
motor at the middle joint).

The second paper \cite{Salge2012} discusses the quasi-linear
Gaussian method as a faster approximation for empowerment, and
focusses on the pendulum; both to compare the QLG method with previous
approximations, and to investigate how different parameters affect the
empowerment map. In the following section we will use the simple
pendulum from the second paper to outline some of the challenges in applying continuous empowerment.

\subsection{Pendulum}

The scenario we will focus on is that of a simple pendulum, because it
incarnates many features typical for the continuous empowerment
scenarios. First we will produce an empowerment map, which assigns an
empowerment value for each state the pendulum can be in. Then we demonstrate empowerment-driven control; an algorithm that generates actions for the pendulum by greedily maximising its expected empowerment in the following step.

We start by observing that the pendulum's current state at the time
$t$ is completely characterized by its angle $\phi$ and its angular
velocity $\dot{\phi}$.

For the model we time-discretize the input. So, the actuator variable
$A$ contains real values $a_t$, which represents the external
acceleration applied to the pendulum. So, at time $t$, the motor
acceleration is set to $a_t$, and this acceleration is then applied
for the duration $\Delta t$. At the end of $\Delta t$, we will
consider the system to be in time $t+\Delta t$, and the next value is
applied.


\subsection{Action Selection}

In general, just having a state-dependent utility function, which
assigns a utility to each state (such as empowerment) does not
immediately provide a control strategy. One way to address this is to
implement a greedy action selection strategy, where each action is
chosen based on the immediate expected gain in empowerment. Note that
empowerment is not a true value function, i.e.\ following its maximum
local gradient does not necessarily correspond to optimizing some
cumulated reward. 

For the \emph{discrete and deterministic} case, implementing a greedy
control is simple. Since we have local model knowledge, we know what
state each action $a$ will lead to. We can then evaluate the empowerment
for each action $a$ that can be taken in the current state, and pick
that action that leads to the subsequent state with the largest
empowerment. This basically provides a gradient ascent approach
(modulated by the effect of the action on the dynamics) on the
empowerment landscape, with all is benefits and drawbacks.

If we are dealing with a \textit{discrete but noisy} system, one needs
to specify in more detail what a ``greedy'' action selection should
look like, since empowerment is not a utility function in the strict
sense of utility theory, and the average empowerment over the
successor states is not the same as the
empowerment of the averaged dynamics. This means that one has
different ways of selecting the desired action for the next step.


However, the most straightforward way remains, of course, the
selection of the highest average empowerment when a particular action
is selected. Assume that, given an action $a$, and a fixed starting
state which we do not denote separately, one has the probability
$p(s|a)$ of getting into a subsequent state $s$\footnote{To simplify the
  argument, we consider here only fully observed states $s$.}.
Each of these successor states $s$ has an associated empowerment value
$\mathfrak{E}(s)$. Thus, the expected
empowerment for carrying out the action $a$ is given by 
\begin{equation}
E[\mathfrak{E}(S)|a]= \sum_{s \in \mathcal{S}}\mathfrak{E}(s)p(s|a)
\label{empinegration}
\end{equation}
and one selects the action with the highest expected empowerment.

The necessity of distinction of deterministic and noisy cases becomes
even more prominent in the continuous case, where the situation is
more complicated. As we have to treat the continuum as a noisy system,
there is usually no unique resulting state for an action $a$, but
rather a continuous distribution density of states $p(s|a)$. Ideally,
one would integrate the empowerment values over this distribution,
similar to Eq.~(\ref{empinegration}), but since empowerment cannot be
expressed as a simple, integrable function, this is not practicable.
One solution is to simply look at the mean of a sampled distribution
over the resulting states and average their empowerment. At this
point, however, no bounds have been derived on how well this value
represents the empowerment values in the distribution of output
states.

The continuity of actions creates another problem. Even if we can
compute the expected empowerment for a given action, then we still
need to select for which actions we want to evaluate their subsequent
empowerment. Again, one possible option is to sample several actions
$a$, distributed in a regular fashion; for example, one could look at
the resulting states for maximal positive acceleration, for no
acceleration at all, and for maximal negative acceleration, and then
select the best. This may miss the action $a$ with the highest
expected empowerment which might fall somewhere between these sample
points. Potential for future work would lie in developing an efficient
method to avoid expensive searches for the highest-valued successive
expected empowerment.

\subsection{Resulting Control}
\label{sec:Control}

\begin{figure}[htp]
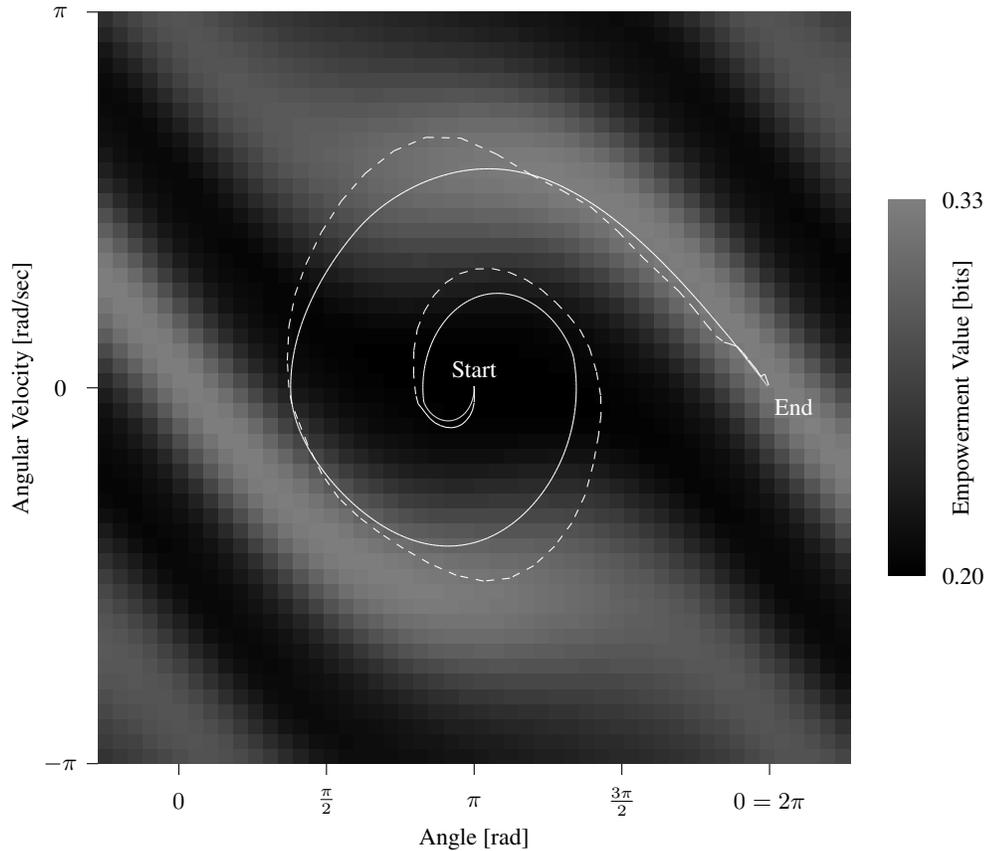


\caption{Graph depicting the state space of a pendulum and its associated empowerment values. The solid line shows the trajectory of a pendulum in this state space, controlled by a greedy empowerment maximization algorithm based on the underlying Gaussian quasilinear empowerment landscape (shown in the background). For comparison of control, the dashed line shows the trajectory created by a greedy maximisation based on a Monte Carlo Gaussian empowerment landscape (not depicted here).}
\label{Control}
\end{figure}

In Fig.~\ref{Control} we can see a empowerment map for the pendulum,
and the resulting trajectory generated by greedy empowerment control.
The controller sampled over 5 possible actuation choices, and chose
the one where the resulting state had the highest expected
empowerment.

In this specific case, the pendulum swings up and comes to rest in the
upright position. This solution, while typical, is not unique. Varying
the parameters for time step length and power constraint can produce
different behaviour, such as cyclic oscillation and resting in the
lower position. We will discuss these cases further below.

One interesting observation to note here is that the empowerment of
the pendulum is not strictly increasing over the run, even though the
control chooses the action that leads to the most empowered successor
state. If one considers the trajectory, it is possible to see that the
pendulum passes through regions where the empowerment lowers again.
This can be seen in Fig.~\ref{Control} where the trajectory passes
through the darker regions of lower empowerment after already being in
much lighter regions of the empowerment map. This is due to the
specific dynamics of the system, in which one can only control the
acceleration of the pendulum, but, of course, not its position change,
which is mediated by the current velocity. So while the controller
chooses the highest empowered future state, all future states have
lower empowerment than the current state.

Contrast this with the discrete maze case: in the latter, the agent could maintain any state of the
environment, i.e. it position, indefinitely, by doing nothing. Greedy
control in the maze therefore moves the agent to increasingly higher
empowered states, until it would reach a local optimum, and then
remain there. 

Strikingly, local empowerment maxima seem to be less of a problem in
the pendulum model (which is, in this respect, very similar to the
mountain-car problem \cite{sutton-barto:rl:1998}). One reason turns
out to be that the pendulum cannot maintain certain positions. If the
pendulum has a non-zero speed, then its next position will be a
different one, because the system cannot maintain both the speed and
position of the pendulum at the same time. This sometimes forces the
pendulum to enter states that are of lower empowerment than its
current state. In the pendulum example this works out well in
traversing the low empowered regions; and the continued local
optimization of empowerment happens to lead to later, even higher
empowered regions.

It is an open question to characterize actuation-perception structures
which would be particularly amenable for the local empowerment
optimization to actually achieve global empowerment optimization or at
least a good approximation of global empowerment optimization. At this
point, it is clear that sharp changes in the empowerment landscape
(e.g.\ discovery of new degrees of freedom, e.g.\ because of the
presence of a new manipulable object) need to be inside the local
exploration range of the action sequences used to compute empowerment.
However, in the case of the pendulum, the maximally empowered point of
the upright pendulum seems to ``radiate'' its basin of attraction into
sufficiently far regions of the state space for the local greedy
optimization to pick this up. The characterization of the properties
that the dynamics of the system needs to have for this to be
successful is a completely open question at this point. Given the
examples studied in \cite{jung2011empowerment}, a cautious hypothesis
may suggest that dynamic scenarios are good candidates for such a
phenomenon.

\subsection{Power Constraint}

\begin{figure}[htp]
\begin{tikzpicture}[scale = 0.85]
\pgftext{\includegraphics[scale=1]{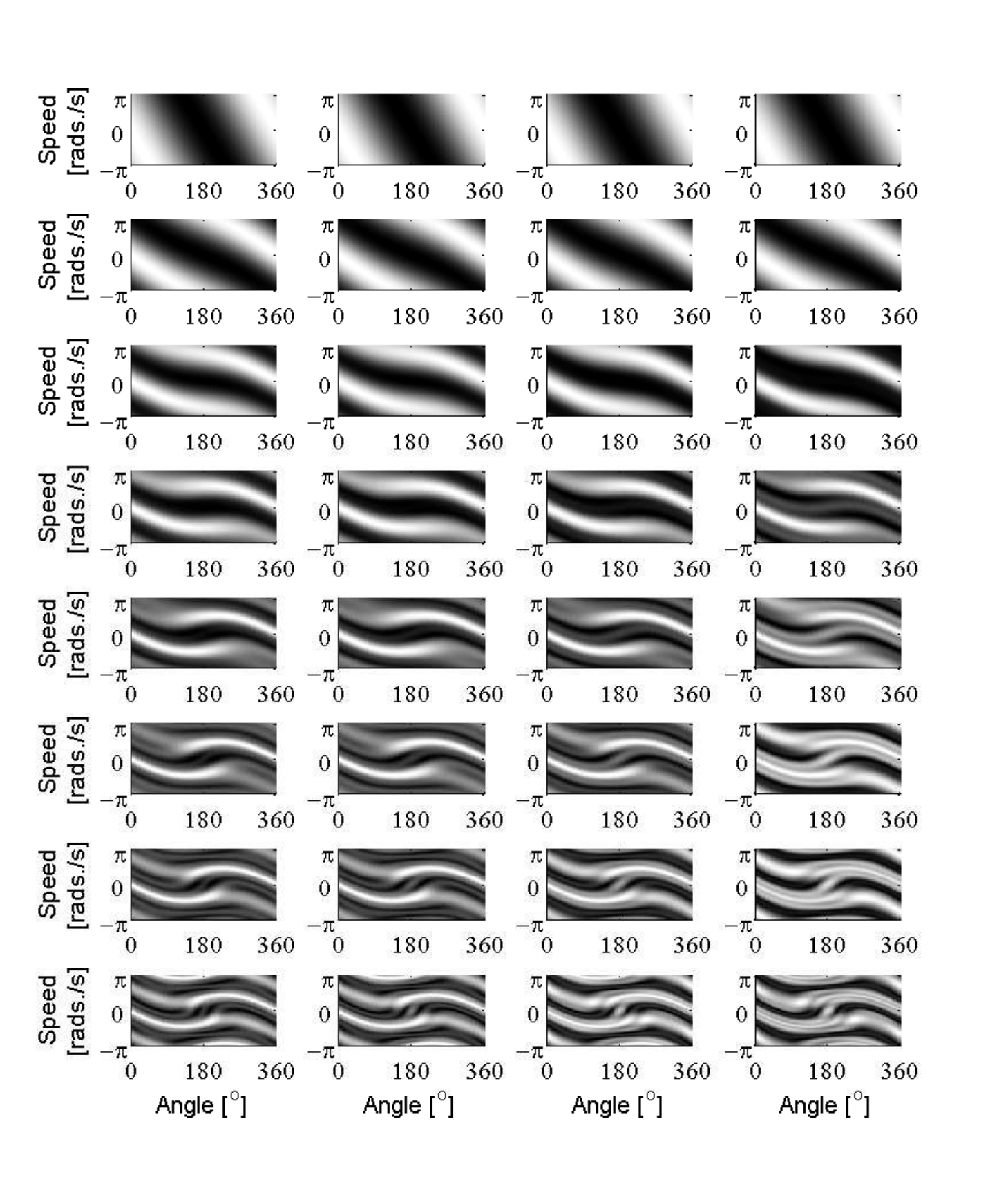}};
\draw[very thick,->,>=triangle 90](-5.5,8.5) -- node[midway,above,scale = 1.5]{Increase in Power (P)} (6,8.5);
\draw (-4.4,8.6) -- (-4.4,8.3) node[below]{0.1};
\draw (-1.4,8.6) -- (-1.4,8.3) node[below]{0.5};
\draw (1.6,8.6) -- (1.6,8.3) node[below]{1};
\draw (4.6,8.6) -- (4.6,8.3) node[below]{5};

\draw[very thick,->,>=triangle 90,xshift = 14.7cm](-7.5,7.5) -- node[midway,below,scale = 1.5,rotate = 90]{Increasing time step ($\Delta t$)} (-7.5,-7.5);

\draw[white, fill = white] (-7,-7.3) rectangle (7,-8);
\node at (-4.4,-7.5) {Angle [$^\circ$]};
\node at (-1.3,-7.5) {Angle [$^\circ$]};
\node at (1.8,-7.5) {Angle [$^\circ$]};
\node at (4.9,-7.5) {Angle [$^\circ$]};

\draw[white, fill = white] (-7,7.5) rectangle (-6,-8);

\foreach \x in {1,...,8}
	\node[rotate = 90] at (-6.5,1.855 * \x  -8.1) {Speed};
\foreach \x in {1,...,8}
	\node[rotate = 90] at (-6.1,1.855 * \x  -8.1) {[rads./s]};
\foreach \x/\xtext in {1/1.5,2/1.3,3/1.1,4/0.9,5/0.7,6/0.5,7/0.3,8/0.1}
	\draw[xshift = 14.7cm]  (-7.4,1.855 * \x  -8.1) --(-7.7,1.855 * \x  -8.1) node[left] {\xtext};
\end{tikzpicture}
\caption{A visualization of the different empowerment landscapes resulting from computation with different parameters for time step length $\Delta t$ and power constraint $P$. The graphs plot empowerment for the two dimensional state space (angular speed, angular position) of the pendulum. White areas indicate the highest empowerment, black areas the lowest possible empowerment. The lower rest position is in the middle of the plots, and has low empowerment for less powered scenarios. The upper rest position is high empowered in all cases, it is located in the middle of the right or left edge of the plots. The areas of high empowerment  close to the upright angel are those were the angular speed moves the pendulum towards the upper rest position. Figure is taken from \cite{Salge2012}}
\label{AllPowerDeltaT}
\end{figure}
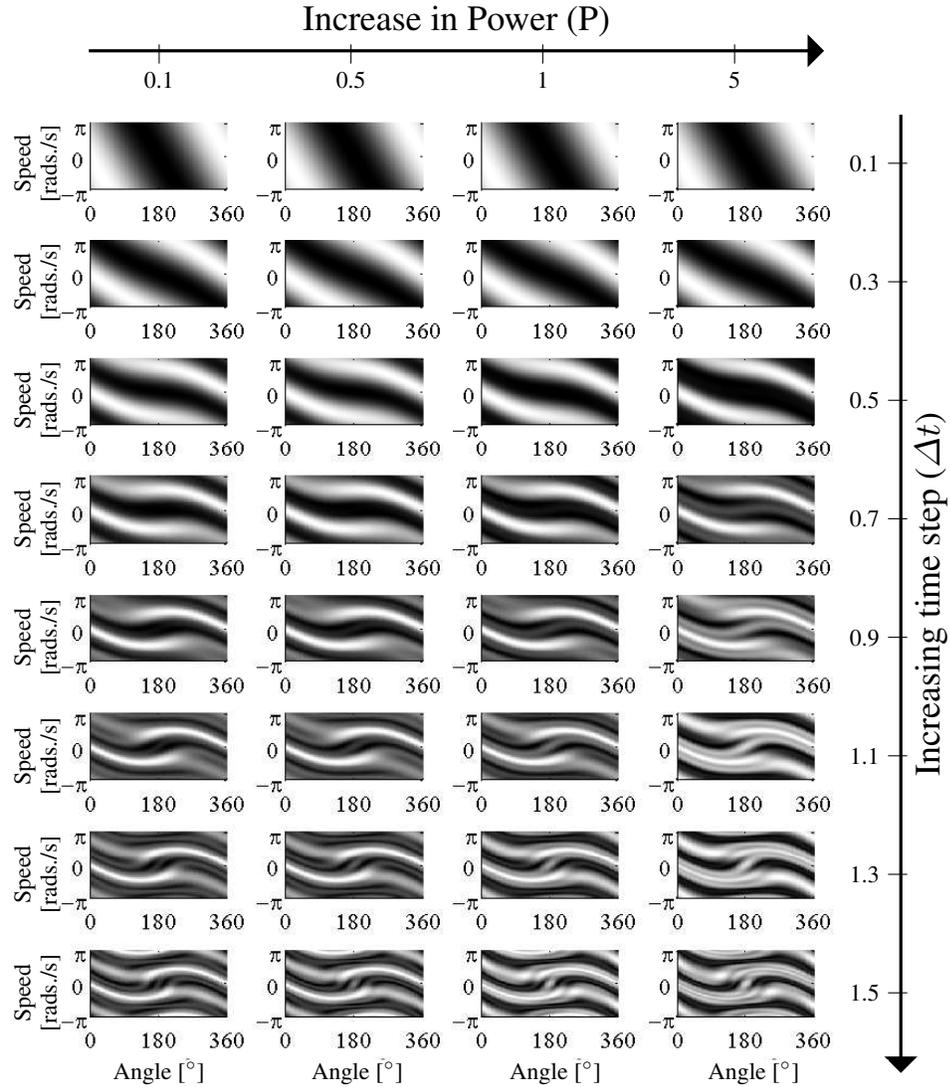

A closer look at the different underlying empowerment landscapes of
the quasi-linear approximation in Fig. \ref{AllPowerDeltaT} shows
their changes in regard to power constraint $P$ and time step length
$\Delta t$.

How the change in the time step duration $\Delta t$ affects the
empowerment, and also how it leads to worse approximations is studied
in greater detail in \cite{Salge2012}. In general, it is not
surprising that empowerment is indeed affected by it, in particular as the
time step duration is closely related to the horizon length. The basic
insight is, however, that a greater time step length allows a further
look-ahead into the future, at the cost of a worsening approximation
with the local linear model.

A more interesting effect in regard to the general applicability of
the fast QLG method is the varying power constraint $P$. In general,
an increase in power will result in an increase in empowerment, no
matter where in the state space the system is. This is not immediately
visible in the figures shown, since the colouring of the graphs is
normalized, so the black and white correspond to the lowest and
highest empowerment value in the respective subgraph. 

A more unexpected effect, however, is a potential \textit{inversion}
of the empowerment landscape as seen in Fig. \ref{AllPowerDeltaT}.
Inversion means that for two specific points in the state space it
might be that for one power level the first has a higher empowerment
than the other, but for a different power level this relationship is
reversed, and now the second point has a higher empowerment. For
example, in Fig. \ref{AllPowerDeltaT} we can consider the row of
landscapes for a $\Delta t$ of 0.7. With increasing power there
appears a new ridge of local maximal empowerment around the lower rest
position of the pendulum.

This slightly counterintuitive effect is a result of how the capacity is distributed on the separate parallel channels. Be reminded, each channel $i$ contributes its own amount to the overall capacity

\begin{equation}
C =\max_{P_i}\sum_i {\frac{1}{2}\log(1 + {\sigma_i P_i}})
\label{capacity2}
\end{equation}
subject to the total power constraint $P$. Depending on the different values for $\sigma_i$, power is first allocated to the channel with the highest amplification value $\sigma_i$, up to a point were the return in capacity for the invested power diminishes so much that adding power to a different channel yields more capacity. From that point on the overall system acts as if it was one channel of bigger capacity. 

In other words, for low power the factor that determines the channel
capacity is the value of the largest $\sigma$ alone. Once the power
increases, the values of both the $\sigma$ become important. It is
therefore possible that for low power, a point with one large $\sigma$
has comparatively high empowerment, while for a higher power level,
another point has a higher empowerment, because the combination of all
the $\sigma$ is better. This is what actually happens in the pendulum
example and causes the pendulum to remain in the lower rest position
in the examples with higher power.

This indicates that the the empowerment-induced dynamics is sensitive
to the given power constraint. One interpretation is that agents with
weak actuators need to fine-tune their dynamics to achieve
high-empowered states. However, agents with strong actuators can
afford to stay in the potential minimum of the system, as their engine
is strong enough to reach all relevant points without complicated
strategems (``if in doubt, use a bigger hammer''). The inversion
phenomenon is a special case for a more generic principle that force
may be used to change the landscape in which the agent finds itself.

Another observation emerging from the inversion phenomenon is the
general question of whether the Gaussian choice for the input
distribution is appropriate. We know that some form of constraint must
be applied, otherwise one could just chose input distributions that
are spaced so far apart that they would fully compensate for the
noise, giving rise to an (unrealistic) infinite channel capacity. Not
only is this unhelpful, but also, as realistic actuations will be
usually limited. In the current model, inspired by well-established
channel capacity applications in communication theory, the power
constraints reflects how limited amount of energies are allotted to
broadcast a signal. But if we instead look, for example, at the
acceleration which a robot could apply to its arm, then for instance
an interval constraint would be much more natural to apply. For
instance, an action $a$ the robot could chose would lie, for example, between
-4.0 and +10.0 $m/s^2$; a servo-based system may, instead specify a
particular location instead, but still constrained by a hard-bounded
interval. As consequence, it might be better to have a model where, instead
of a general power constraint $P$, a hard upper and lower limit for
each dimension of the actuator input $A$ is imposed. At present, we are not aware
of a method to directly compute the channel capacity for a multiple
input, multiple output channel with coloured Gaussian noise that uses
such a constraint.

\subsection{Model Acquisition}

Before we end this overview we will at least briefly address the
problem of model acquisition or model learning. As mentioned, empowerment needs the
model $p(s|a,r)$ for its computation. Strictly spoken, the acquisition or adaptation
of this model is not part of the empowerment formalism. It is external
to it, the model being either given in advance, or being acquired by
one of many candidate techniques. However, given that empowerment will
be used in scenarios where the model is not known and has to be learnt
at the same time as the empowerment gradient is to be followed, model
acquisition needs to be treated alongside the empowerment formalism
itself. 

As mentioned, empowerment only needs a local model of the dynamics
from the agent's actuators to the agent's sensor in the current state
of the world, but this local model is essential to compute
empowerment.

Much of the earlier empowerment work operates under the assumption
that the agent in question has somehow obtained or is given a
sufficiently accurate model $p(s|a,r)$. Without addressing the
``how'', this acknowledges the fact that an agent-centric, intrinsic
motivation mechanism needs to have this forward model available within
the agent.

The earliest work to touch on this \cite{klyubin2008keep} deals with
context-dependent empowerment. To model the relationship between an
AIBO's discrete actions, and some discrete camera inputs, regular
motions of the head are performed to sample the environment. These
were then used to construct joint probability distributions and select
an appropriate separation of all states of $R$ into different
contexts. The choice of context itself was also a decision on how to
internally represent the world in an internal model, especially if
there is only limited ``resources'' available to model the world. By
grouping together states that behave similarly, the agent gets a good
approximation of the world dynamics, and its internal empowerment
computation results in high-empowered states. If the agent groups
states with different behaviour together, then the resulting contexts
have higher levels of uncertainty, and result in comparatively lower
empowerment values (from the agent's perspective).

In general, it is clear that the quality of the model will affect the
internal evaluation of empowerment. If the dynamics of a state are
modelled with a great degree of uncertainty, then this noise will also
reflect negatively in the empowerment value for this state. The
interesting question here is then how to distinguish between those
states that are truly random, and those where the action model is just
currently not well known. This also indicates another field of future
research. The hypothesis is that, if we would model how exploration or
learning would affect our internal model, then the maximisation of
(internally computed) empowerment could also lead to exploration and
learning behaviour.

In the continuous case, we have to deal with the additional question
on how to best represent the conditional probability distributions,
since, unlike the discrete case, there is no general and exact way of
doing so. \citet{jung2011empowerment} uses Gaussian Processes to store
the dynamics of the world. This also offers a good interface between the
use of a Gaussian Process Learner and the Monte Carlo integration with
assumed Gaussian noise. The faster quasi-linear Gaussian approximation
\cite{Salge2012} also interact well with representation, and,
conveniently, the covariance metric used for the coloured noise can be
directly derived from the GP. In general, one would assume that other
methods and algorithms to acquire a world model could be similarly
combined with empowerment. It remains an open question which of these
models are well suited, not just as approximations of the world
dynamics in general, but in regard to how well they represent those
aspects of the world dynamics that are relevant to attain high
empowerment values.

\section{Conclusion}

The different scenarios presented here, and in the literature
on empowerment in general, are highlighting an important aspect of the
empowerment flavour of intrinsic motivation algorithms, namely its
\textit{universality}. The same principle that organizes a swarm of
agents into a pattern can also swing the pendulum into an upright
position, seek out a central location in a maze, be driven towards a
manipulable object, or drive the evolution of sensors.

The task-independent nature reflected in this list can be both a
blessing and a curse. In many cases the resulting solution, such as
swinging the pendulum into the upright position, is the goal implied
by default by a human observer. However, if indeed a goal is desired
that differs from this default, then empowerment will not be the best
solution. At present, the question of how to integrate explicit
non-default goals into empowerment is fully open.

Another strong assumption that comes with the use of empowerment is
its local character. On the upside, it simplifies the computation and
makes the associated model acquisition much cheaper as only a very
small part of the state space ever needs to be explored; the
assumption of the usefulness of empowerment as a proxy principle for
other implicit and less accessible optimization principles depends
heavily on how well the local structure of the system dynamics will
reflect its global structure. The precise nature of this phenomenon is
not fully understood in the successful scenarios, but is believed to
have to do with the regularity (e.g.\ continuity/smoothness) of the
system dynamics. Of course, if any qualitative changes in the dynamics
happen just outside of the empowerment horizon, the locality of
empowerment will prevent them from being seen. This could be due to some disastrous
``cliff'', or something harmless like the discovery of an object that
can be manipulated. Once, however, the change enters the empowerment
horizon, and assuming that one can obtain a model of how it will
affect the dynamics without losing the agent, empowerment will provide
the gradients appropriate to the change.

Another central problem that, in the past, has reappeared across
different applications is the computational feasibility. Empowerment
quickly becomes infeasible to compute, which is a problem for both the
behavioural empowerment hypothesis, and the application of empowerment
to real-time AI or robotics problems. Newer methods address both the
case for continuous empowerment (such as the QLG), and deeper
empowerment horizons (such as the ``impoverished'' versions of
empowerment). They, of course, come with additional assumptions and
parameters, and provide only approximate solutions, but maintain the
general character of the full solutions, allowing to export
empowerment-like characteristics into domains that were hitherto
inaccessible.

Let us conclude with a remark regarding the biological empowerment
hypotheses in general: the fact that the default behaviours produced
by empowerment seem often to match what intuitive expectations
concerning default behaviour seem to imply, there is some relevance in
investigating whether some of these behaviours are indeed
approximating default behaviours observed in nature. A number of
arguments in favour of why empowerment maximizing or similar behaviour
could be relevant in biology have been made in \cite{klyubin2008keep},
of which in this review we mainly highlighted its role as a measure of
sensorimotor efficiency and the advantages that an evolutionary
process would confer to more informationally efficient
perception-action configurations. 

Together with other intrinsic motivation measures, empowerment is thus
a candidate measure which may help bridge the gap between
understanding how organisms may be able to carry out default
adaptations into their niche in an effective manner, and methods which
would also allow artificial devices to try and copy the success that
biological organisms have in doing so.

\bibliographystyle{apalike}

\bibliography{ref1}

\end{document}